\documentclass[lettersize,journal]{IEEEtran}
\usepackage{amsmath}
\usepackage[utf8]{inputenc} 
\usepackage[T1]{fontenc}    
\usepackage{hyperref}       
\hypersetup{draft}
\usepackage{url}            
\usepackage{booktabs}       
\usepackage{bbm}
\usepackage{amsfonts}       
\usepackage{nicefrac}  
\usepackage{amsthm}
\usepackage{microtype}      
\usepackage{xcolor}         
\usepackage{amsmath}
\usepackage{mathtools}
\usepackage{booktabs}
\usepackage{caption}
\usepackage[ruled,vlined]{algorithm2e}
\usepackage{graphicx}
\DeclareMathOperator*{\argmin}{arg\,min}
\DeclareMathOperator*{\argmax}{arg\,max}
\newtheorem{theorem}{Theorem}

\graphicspath{ {./} }
\usepackage{stackengine}
\usepackage{subcaption}
\usepackage{caption}
\usepackage{graphicx}
\usepackage{float}
\usepackage{diagbox}
\usepackage{makecell}

\graphicspath{ {./} }

\newcommand{\norm}[1]{\left\lVert#1\right\rVert}
\def\BibTeX{{\rm B\kern-.05em{\sc i\kern-.025em b}\kern-.08em
    T\kern-.1667em\lower.7ex\hbox{E}\kern-.125emX}}

\newcolumntype{P}[1]{>{\centering\arraybackslash}p{#1}}
\DeclareMathOperator{\E}{\mathbb{E}}

\ifCLASSINFOpdf
\else
\fi
\usepackage{algorithmic}
\begin{document}
%
\title{User-Centric Federated Learning: Trading off Wireless Resources for Personalization}
%
%
%


\author{Mohamad~Mestoukirdi$^{\dagger}$,~\IEEEmembership{Student Fellow,~IEEE,}
        Matteo Zecchin$^{\dagger}$,~\IEEEmembership{Student Fellow,~IEEE,}
       David~Gesbert,~\IEEEmembership{Fellow,~IEEE,}
        and Qianrui~Li,~\IEEEmembership{Member,~IEEE}
        
 \thanks{$^\dagger$ Equal contribution.}
\thanks{ M. Mestoukirdi, M.  Zecchin, and D. Gesbert are with the Communication
Systems Department, EURECOM, Sophia-Antipolis, France. Emails: \{Mestouki, Zecchin, Gesbert\}@eurecom.fr. M. Mestoukirdi and Q. Li are with Mitsubishi Electric R\&D Centre Europe. Emails: \{M.Mestoukirdi, Q.Li\}@fr.merce.mee.com.}
\thanks{The work of M. Zecchin is funded by the Marie Sklodowska Curie action WINDMILL (grant No. 813999).}
}

\maketitle

\begin{abstract}
Statistical heterogeneity across clients in a Federated Learning (FL) system increases the algorithm convergence time and reduces the generalization performance, resulting in a large communication overhead in return for a poor model. To tackle the above problems without violating the privacy constraints that FL imposes, personalized FL methods have to couple statistically similar clients without directly accessing their data in order to guarantee a privacy-preserving transfer. In this work, we design user-centric aggregation rules at the parameter server (PS) that are based on readily available gradient information and are capable of producing personalized models for each FL client. The proposed aggregation rules are inspired by an upper bound of the weighted aggregate empirical risk minimizer. Secondly, we derive a communication-efficient variant based on user clustering which greatly enhances its applicability to communication-constrained systems. Our algorithm outperforms popular personalized FL baselines in terms of average accuracy, worst node performance, and training communication overhead.
\end{abstract}
\section{Introduction}
In recent years, the evolution of energy and computation-efficient hardware, together with the wide adoption of data-driven solutions have led to an increased interest in pushing intelligence closer to edge devices where data is generated. This contributed to the emergence of autonomous and intelligent systems where decisions are made locally. However, the reliable operation of intelligent edge devices requires periodic ML training and tuning, which mainly relies on pooling data from a multitude of devices toward a central entity. Consequently, privacy concerns arise in such settings, as data owners may be reluctant in sharing sensible and personal pieces of information \cite{health}. Additionally, the soaring model complexity of modern ML solutions requires vast amounts of data to be harvested to achieve satisfactory inference accuracy. This introduces a large communication overhead and long training delays.

Federated Learning (FL) \cite{mcmahan2017communication} was introduced to deal with these problems. It offers clients the possibility of collaboratively training models under the orchestration of a parameter server (PS), by iteratively aggregating locally optimized models without the need to offload any raw data centrally. Such an approach tackles both the privacy and communications challenges mentioned above. Early FL algorithms were devised under the assumption that the data distribution of clients' data sets is common. In this case, clients are said to share the same learning task, and traditional FL (e.g FedAvg \cite{mcmahan2017communication}) algorithms can perform and generalize well yielding a single model, fitting the common data distribution. However, this assumption is hardly met in practice \cite{sattler2020clustered}, as data distribution heterogeneity often arises in distributedly generated data sets. In such cases, traditional FL (e.g FedAvg) approaches exhibit slow convergence and often fail to generalize well \cite{li2018federated}, especially when conflicting objectives among users exist. This is a direct consequence of the fact that in heterogeneous settings, a convex combination of locally trained models may not be fit for any particular client data distribution. Hence, heterogeneous distributions bring about an interesting trade-off: On the one hand the advantage of exploiting training data at other clients when the local training data is insufficient, and on the other hand the problem of having the trained model steered towards improper directions due to differences in data distributions among clients. This trade-off motivates the search for new FL strategies that can navigate the compromise between model aggregation benefits and the threat of model mismatch.

 In \cite{prevwork}, we proposed a novel user-centric aggregation rule to tackle the underlying heterogeneity among clients and overcome the shortcomings of the traditional FL schemes. The proposed strategy leverages user-centric aggregation rules at the PS to produce models at each device that are tailored to their local data distribution. This is achieved by generalizing the aggregation rule introduced by McMahan et al. \cite{mcmahan2017communication}. In the case of a set of $m$ collaborating devices, the original objective in \cite{mcmahan2017communication} produces a common model at each communication round $t$ according to
\begin{equation}
\begin{aligned}
\theta^{t}\leftarrow \sum_{i=1}^m w_i\theta_i^{t-1/2},
\end{aligned}
\label{eq0}
\end{equation}
where each $w_i$ weights the contribution of the locally optimized model $\theta_i^{t-\frac{1}{2}}$ of user $i$, to the update global model $\theta^t$. 
On the other hand, the proposed aggregation rule replaces the weighting coefficients $\{w_i\}^m_{i=1}$ by user-specific weighting vectors $\vec{w}_i = (w_{i,1},\dots,w_{i,m})$ and it produces a personalized model update for each FL client
\begin{equation}
\begin{aligned}
\theta^{{t}}_i \leftarrow \sum^m_{j=1}w_{i,j}\theta^{{t-1/2}}_j \hspace{1cm} \text{for } i = 1,2,\cdots,m
\end{aligned}
\label{eq2}
\end{equation}
The key motivation underpinning the use of distinct user-centric personalization rules is that a single model often fails in heterogeneous settings \cite{mcmahan2017communication}. At the same time, hard clustering strategies \cite{sattler2020clustered,briggs2020federated} are limited to restrictive intra-cluster collaboration and they cannot exploit similarities among different clusters. The authors in \cite{zhang2020personalized} proposed FedFomo, a personalization scheme that uses a similar aggregation policy as ours \cite{prevwork}. However, FedFomo's weighting scheme is repeatedly refined during training and it relies on sharing local models among clients at each communication round. This strategy can violate the FL privacy-preserving nature and introduces a large communication burden to the training procedure. In contrast, our personalization policy is shown experimentally to enjoy faster convergence, being able to capture the data heterogeneity at the start of training without the need for further refinements at later stages.

In this work, we extend the findings of \cite{prevwork}. We derive an upper bound on the expected risk endured by the minimizer of the weighted empirical aggregate loss. Then, we motivate the use of heuristically defined weights in place of the theoretically optimal ones. Furthermore, to limit the communication costs induced by transmitting multiple personalized models, we propose a $K$-means clustering algorithm over the user-centric weights to limit the number of personalized streams, while taking into account the underlying heterogeneous target tasks and highlighting inter-cluster collaboration. This enables a trade-off between the learning accuracy and the communication load in some heterogeneous settings. Finally, we show that the silhouette score over the returned $K$-means solution can detect the underlying heterogeneity, and provides a principled way to choose the number of user-centric rules. Through extensive numerical experiments on FL benchmarks, we demonstrate the performance of our proposed strategy compared to other state-of-the-art solutions, in terms of inference accuracy, and communication costs.

\section{Related Work}
Several recent studies investigate the challenges that arise due to the underlying task heterogeneity present across learners in Federated Learning settings. For instance, \cite{briggs2020federated,sattler2020clustered} devised a hierarchical clustering scheme to group users that share the same learning task and enable collaboration among them only. However, their strategy is based on the assumption that heterogeneous tasks are either tangential or parallel, which is not necessarily true, as tasks are defined by the users' target data distributions which are often different for each of them. In this sense, hard-clustering strategies limit the degree of collaboration across learners and may not always be able to capture the differences across users' tasks. In \cite{mixture2021s} a distributed Expectation-Maximization (EM) algorithm has been proposed, that concurrently converges to a set of shared hypotheses and a personalized linear combination of them at each device. Similarly in \cite{reisser2021federated}, a  Mixture of  Experts' formulation has been devised to learn a  personalized mixture of the outputs of a jointly trained set of models. Similar to \cite{zhang2020personalized}, exploiting the full personalization potential of the solutions in \cite{mixture2021s,reisser2021federated} induces a huge overhead over the communication resources in the federated system, which renders their approaches unpractical. Similar to Fedprox \cite{DBLP:journals/corr/abs-1812-06127}, the authors in \cite{DBLP:journals/corr/abs-1910-06378} propose SCAFFOLD to tackle the  ``\textit{client drifts}" that emerge as a result of the heterogeneity of the clients' data sets during the global model training. However, in some heterogeneous settings, ``\textit{client drifts}" can act as an indication of the existence of opposing target tasks among the learners. Therefore, intelligently employing the drifts can highlight similarity patterns among the clients' tasks \cite{sattler2020clustered}, which in turn can aid in training multiple refined models to fit each of the available tasks, yielding better personalized models in contrast to a single global model trained by SCAFFOLD. More recently, the authors in \cite{DBLP:journals/corr/abs-2012-04221} propose Ditto, where users collaborate to train a separate global model akin to \cite{mcmahan2017communication}, which is then used to steer the training of the local personalized model at each user via local model adaptation. Their approach embodies the intuition of pFedMe \cite{DBLP:journals/corr/abs-2006-08848}, which decouples personalized model optimization from the global model learning by introducing a penalizing term to regularize the clients local adaptation step. Despite resulting in a per-user personalized model, collaboration among users in Ditto and pFedMe is limited to updating the global model, while relying solely on the local data sets to train their personalized models, rather than leveraging collaboration among statistically similar learners to refine those models. Consequently, the resulting personalized models may generalize poorly, especially in settings where local data sets are small in size.

\label{section2}

\section{Learning with heterogeneous data sources}
\label{section3}
\label{sec3}
In this section, we provide theoretical guarantees for learners that combine data from heterogeneous data distributions. The set-up mirrors the one of personalized federated learning and the results are instrumental to derive our user-centric aggregation rule. In the following, we limit our analysis to the discrepancy distance, but it can be readily extended to other divergences as we show later.

In the federated learning setting, the weighted combination of the empirical loss terms of the collaborating devices represents the customary training objective. Namely, in a distributed system with $m$ nodes, each endowed with a data set $\mathcal{D}_i$ of $n_i$ IID samples from a local distribution $P_i$, the goal is to find a predictor $f:\mathcal{X}\to\mathcal{\hat{Y}}$ from a hypothesis class $\mathcal{F}$ that minimizes
\begin{equation}
    L(f,\vec{w})=\sum_{i=1}^m \frac{w_i}{n_i}\sum_{(x,y)\in \mathcal{D}_i}\ell(f(x),y)
    \label{aggregatedloss}
\end{equation}
where  $\ell: \mathcal{\hat{Y}}\times \mathcal{Y}\to \mathbb{R}^+$ is a loss function and $\vec{w}=(w_1,\dots,w_m)$ is a weighting scheme. In case of identically distributed local data sets, the typical weighting vector is $\vec{w}=\frac{1}{\sum_i n_i}\left(n_1,\dots,n_m\right)$, the relative fraction of data points stored at each device. This particular choice minimizes the variance of the aggregated empirical risk, which is also an unbiased estimate of the local risk at each node in this scenario. However, in the case of heterogeneous local distributions, the minimizer of $\vec{w}$-weighted risk may transfer poorly to certain devices whose target distribution differs from the mixture $P_{\vec{w}}=\sum^m_{i=1}w_iP_i$. Furthermore, it may not exist a single weighting strategy that yields a universal predictor with satisfactory performance for all participating devices. To address the above limitation of a universal model, personalized federated learning allows adapting the learned solution at each device. In order to better understand the potential benefits and drawbacks coming from the collaboration with statistically similar but not identical devices, let us consider the point of view of a generic node $i$ that has the freedom of choosing the degree of collaboration with the other devices in the distributed system. Namely, identifying the degree of collaboration between node $i$ and the rest of users by the weighting vector $\vec{w}_i=(w_{i,1},\dots,w_{i,m})$ (where $w_{i,j}$ defines how much node $i$ relies on data from user $j$) we define the personalized objective for user $i$ 
\begin{equation}
   \min_{\theta} L(\theta)=\sum_{i=1}^m \frac{w_{i}}{|\mathcal{D}_i|}\sum_{(x,y)\in \mathcal{D}_i}\ell(f_{\theta}(x),y)
\label{centricloss}
\end{equation}
and the resulting personalized model 
\begin{equation}
    \hat{f}_{\vec{w}_i}=\argmin_{f\in\mathcal{F}}L(f,\vec{w}_i).
    \label{learner}
\end{equation}
We now seek an answer to: \emph{``What's the proper choice of $\vec{w}_i$ in order to obtain a personalized model $\hat{f}_{\vec{w}_i}$ that performs well on the target distribution $P_i$?''}. This question is deeply tied to the problem of domain adaptation, in which the goal is to successfully aggregate multiple data sources in order to produce a model that transfers positively to a different and possibly unknown target domain. In our context, the data set $\mathcal{D}_i$ is made of data points drawn from the target distribution $P_i$ and the other devices' data sets provide samples from the sources $\{P_j\}_{j\neq i}$.  Leveraging results from domain adaptation theory \cite{ben2010theory}, we provide learning guarantees on the performance of the personalized model $\hat{f}_{\vec{w}_i}$ to gauge the effect of collaboration that we later use to devise the weights for the user-centric aggregation rules.\newline\newline In order to avoid negative transfer, it is crucial to upper bound the performance of the predictor w.r.t. to the target task. The discrepancy distance introduced in \cite{mansour2009domain} provides a measure of similarity between learning tasks that can be used to this end. For a hypothesis set of functions $\mathcal{F}:\mathcal{X}\rightarrow\hat{\mathcal{Y}}$ and two distributions $P,Q$ on $\mathcal{X}$, the discrepancy distance is defined as 
\begin{equation}
    d_{\mathcal{F}}(P,Q)=\sup_{f,f'\in \mathcal{F}}\left|\mathbb{E}_{x\sim P}\left[\ell(f,f')\right]-\mathbb{E}_{x\sim Q}\left[\ell(f,f')\right]\right|
     \label{disc_dist}
\end{equation}
where we streamlined notation denoting $f(x)$ by $f$.
For bounded and symmetric loss functions that satisfy the triangular inequality,  the previous quantity allows to obtain the following inequality
\begin{equation*}
    \mathbb{E}_{(x,y)\sim P}[\ell(f,y)]\leq \mathbb{E}_{(x,y)\sim Q}[\ell(f,y)]+d_{\mathcal{F}}(P,Q)+\gamma
\end{equation*}
where $\gamma=\inf_{f \in \mathcal{F}}\left(\mathbb{E}_{(x,y)\sim P}[\ell(f,y)]+\mathbb{E}_{(x,y)\sim Q}[\ell(f,y)]\right)$. We can exploit the inequality to obtain the following risk guarantee for $\hat{f}_{\vec{w}_i}$ w.r.t the true minimizer $f^*$ of the risk for the distribution $P_i$.

\begin{theorem}
For a loss function $\ell$ $B$-bounded range, symmetric and satisfying the triangular inequality, with probability $1-\delta$ the function $f_{\vec{w_i}}$  satisfies
\begin{align*}
    &E_{z\sim P_i}[\ell(\hat{f}_{\vec{w}_i},z)]-E_{z\sim P_i}[\ell(f^*,z)]\leq \\&B \sqrt{\sum^m_{j=1}\frac{w^2_{i,j}}{n_j}}\left(\sqrt{\frac{2d}{\sum_i n_i}\log\left(\frac{e\sum_i n_i}{d}\right)}+\sqrt{\log\left(\frac{2}{\delta}\right)}\right)+\\&2 \sum_{j=1}^m w_{i,j}d_{\mathcal{F}}(P_i,P_{j})+2\gamma
\end{align*}
where $\gamma=\min_{f\in \mathcal{F}}\left( E_{z\sim P_i}[\ell(f,z)]+E_{z\sim P_{\vec{w}_i}}[\ell(f,z)]\right)$ and $d$ is the VC-dimension of the function space resulting from the composition of $\mathcal{F}$ and $\ell$.
\label{Th1}
\end{theorem}
Recently, an alternative bound based on an information theoretic notion of dissimilarity, the Jensen-Shannon divergence, has been proposed \cite{shui2020beyond}. It is based on less restrictive constraints, as it only requires the loss function $\ell(f,Z)$ to be sub-Gaussian of some parameter $\sigma$ for all $f\in \mathcal{F}$, and therefore whenever $\ell(\cdot)$ is bounded, the requirement is automatically satisfied. Measuring similarity by the Jensens-Shannon divergence the following inequality is available
\begin{equation}
    E_{X\sim P}[X]\leq E_{X\sim Q}[X] +\beta\sigma^2+\frac{D_{JS}(P||Q)}{\beta} \quad \text{for $\beta>0$}
    \label{JS}
\end{equation}

                                                                                                                                     where $D_{JS}(P\|Q)=\text{KL}\left(P\Big\|\frac{P+Q}{2}\right)+\text{KL}\left(Q\Big\|\frac{P+Q}{2}\right)$.
Exploiting the above inequality we obtain the following estimation error bound.
\begin{theorem}
For a loss function $\ell$ $B$-bounded range, the function $f_{\vec{w_i}}$  satisfies
\begin{align*}
    &E_{z\sim P_i}[\ell(\hat{f}_{\vec{w}_i},z)]-E_{z\sim P_i}[\ell(f^*,z)]\leq \\ &B \sqrt{\sum^m_{j=1}\frac{w^2_{i,j}}{n_j}}\left(\sqrt{\frac{2d}{\sum_i n_i}\log\left(\frac{e\sum_i n_i}{d}\right)}+\sqrt{\log\left(\frac{2}{\delta}\right)}\right)+\\&B\sqrt{2\sum^m_{j=1}w_{i,j}D_{JS}(P_i||P_j)}
\end{align*}

\label{Th2}
\end{theorem}
\textbf{Proof of Theorem \ref{Th1} and \ref{Th2}:} In the Appendix \ref{A}.\newline
The theorems highlights that a fruitful collaboration should strike a balance between the bias terms due to dissimilarity between local distribution and the risk estimation gains provided by the data points of other nodes. Minimizing the upper bound in Th. \ref{Th1},\ref{Th2} with respect to the user-specific weights, and using the optimal weights in our aggregation rule seems an appealing solution to tackle the data heterogeneity during training; however, the distance terms $\left(d_{\mathcal{F}}(P_i,P_k) \text{ and }D_{JS}(P_i||P_j)\right)$ are difficult to compute, especially under the privacy constraints that federated learning imposes. For this reason, in the following we consider a heuristic method based on the similarity of the readily available users' model updates to estimate the collaboration coefficients.

\section{User-centric aggregation}
\label{section4}
For a suitable hypothesis class parametrized by $\theta\in \mathbb{R}^d$, federated learning approaches use an iterative procedure to minimize the aggregate loss (\ref{aggregatedloss}) with $\vec{w}=\frac{1}{\sum_i n_i}\left(n_1,\dots,n_m\right)$. At each round $t$, the PS broadcasts the parameter vector $\theta^{t-1}$ and then combines the locally optimized models by the clients $\{\theta_i^{t-1}\}_{i=1}^m$ according to the following aggregation rule
\[
\theta^{t}\leftarrow \sum_{i=1}^m\frac{n_i}{\sum_{j=1}^m n_j}\theta_i^{t-1}.
\]
As mentioned in Sec. \ref{sec3}, this aggregation rule has two shortcomings: it does not take into account the data heterogeneity across users, and it is bounded to produce a single solution. For this reason, we propose a user-centric model aggregation scheme that takes into account the data heterogeneity across the different nodes participating in training and aims at neutralizing the bias induced by a universal model. Our proposal generalizes the naïve aggregation of FedAvg, by assigning a unique set of mixing coefficients $\vec{w}_i$ to each user $i$ and, consequently, a user-specific model aggregation at the PS side. Namely, on the PS side, the following set of user-centric aggregation steps are performed
\begin{equation}
\begin{aligned}
\theta^{{t}}_i \leftarrow \sum^m_{j=1}w_{i,j}\theta^{{t-1/2}}_j \quad \textnormal{for $i=1,\dots,m$}
\end{aligned}
\label{eq3}
\end{equation}
 where now, $\theta^{{t-1/2}}_j$ is the locally optimized model at node $j$ starting from $\theta^{{t-1}}_j$, and $\theta^{{t}}_i$ is the user-centric aggregated model for user $i$ at communication round $t$.
\begin{figure}[h!]
         \centering
         \includegraphics[width=0.335\textwidth]{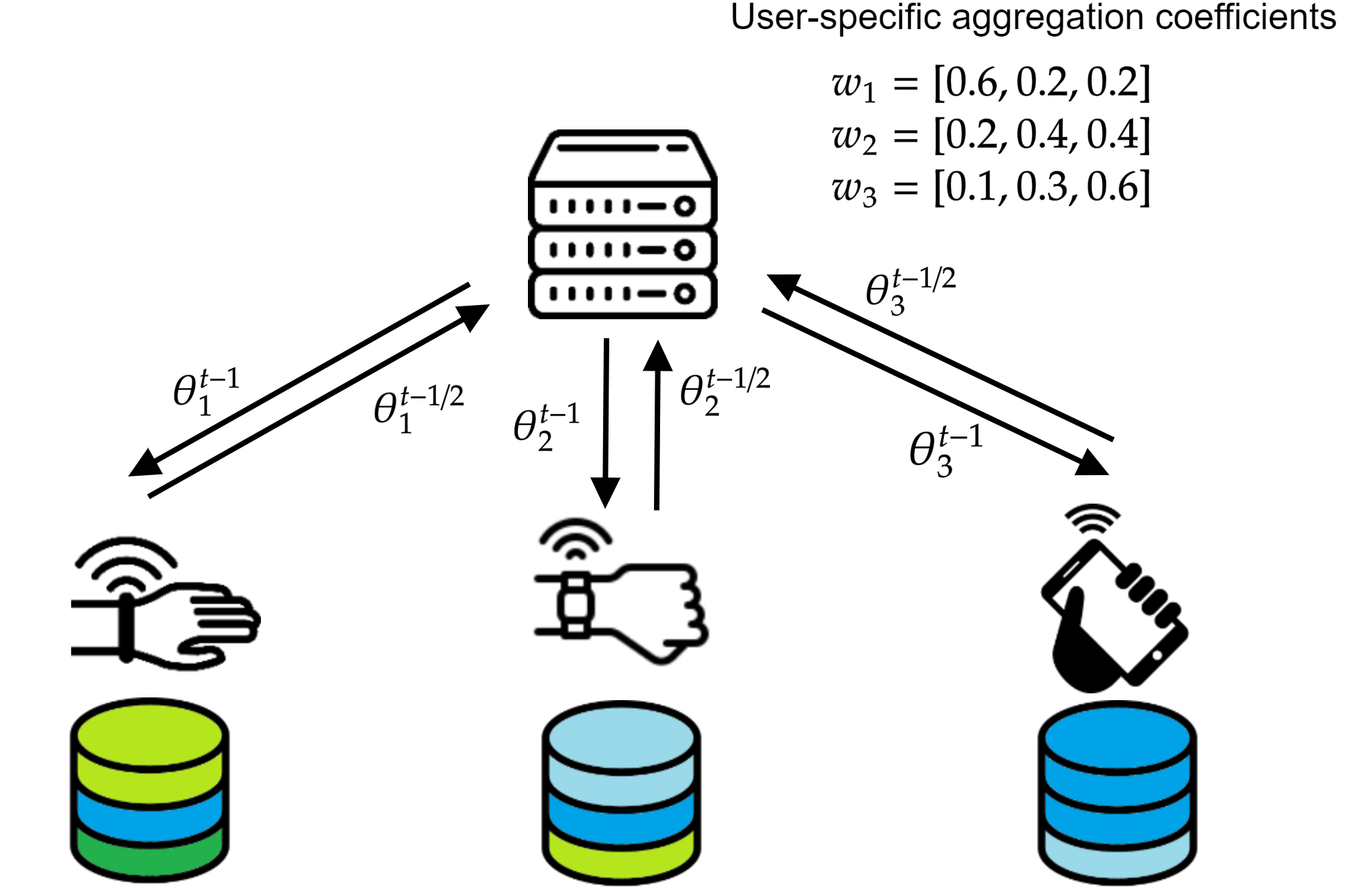}
         \label{fig:system_model}
     \caption{Personalized Federated Learning with user-centric aggregates at round $t$.}
     
\end{figure}\
As we elaborate next, the mixing coefficients are heuristically defined based on a distribution similarity metric and the data set size ratios. These coefficients are calculated before the start of federated training. The similarity score we propose is designed to favour collaboration among similar users and takes into account the relative data set sizes, as more intelligence can be harvested from clients with larger data availability. Using these user-centric aggregation rules, each node ends up with its personalized model that yields better generalization for the local data distribution. It is worth noting that the user-centric aggregation rule does not produce a minimizer of the user-centric aggregate loss given by (\ref{centricloss}). At each round, the PS aggregates model updates are computed starting from a different set of parameters. Nonetheless, we find it to be a good approximation of the true update since personalized models for similar data sources tend to propagate in a close neighbourhood. The aggregation in \cite{zhang2020personalized} capitalizes on the same intuition.
\subsection{Computing the Collaboration Coefficients}
\label{collaboration}
Computing the discrepancy distance (\ref{disc_dist}) can be challenging in high-dimension, especially under the communication and privacy constraints imposed by federated learning. For this reason, we propose to compute the mixing coefficient based on the relative data set sizes and the distribution similarity metric given by
\begin{align*}
\Delta_{i,j}(\hat{\theta}) = & \norm{ \frac{1}{n_i}\sum_{(x,y)\in \mathcal{D}_i}{\hspace{-10pt}\nabla \ell(f_{\hat{\theta}},y)} - \frac{1}{n_j}\sum_{(x,y)\in \mathcal{D}_j}\hspace{-10pt}\nabla \ell(f_{\hat{\theta}},y) }^2\\
\approx & \norm{ \mathbb{E}_{z\sim P_i}\nabla \ell(f_{\hat{\theta}},y) - \mathbb{E}_{z\sim P_j}\nabla \ell(f_{\hat{\theta}},y) }^2
\label{eq3}
\end{align*}
where the quality of the approximation depends on the number of samples $n_i$ and $n_j$.  The mixing coefficients for user $i$ are then set to the following normalized Gaussian kernel function
\begin{equation}
  w_{i,j}=\frac{\frac{n_j}{n_i}e^{-\frac{1}{2\sigma_i\sigma_j}\Delta_{i,j}(\hat{\theta}) }}{\sum^m_{j'=1}\frac{n_{j'}}{n_i}e^{-\frac{1}{2\sigma_i\sigma_j}\Delta_{i,j'}(\hat{\theta}) }} \hspace{1cm}\textnormal{for $j=1,\dots,m$}
  \label{eq7}
\end{equation}
 The mixture coefficients are calculated at the PS during a special round before federated training. During this round, the PS broadcasts an initialized model denoted ($\hat{\theta}$ = $\theta^0$) to the users, which computes the full gradient on their local data sets. At the same time, each node $i$ locally estimates the value $\sigma^2_i$ partitioning the local data randomly in $K$ batches $\{\mathcal{D}^k_i\}^K_{k=1}$ of size $n_k$ and computing
\begin{equation} 
\sigma^2_i = \frac{1}{K}\sum_{k=1}^K\norm{\frac{1}{n_k} \sum_{(x,y)\in \mathcal{D}^k_i}\hspace{-10pt}\nabla \ell(f_{\hat{\theta}},y)-\frac{1}{n_i}\sum_{(x,y)\in \mathcal{D}_i}\hspace{-10pt}\nabla \ell(f_{\hat{\theta}},y )}^2   \label{eq9}
\end{equation} 
where $\sigma^2_i$ is an estimate of the gradient variance (i.e noise) computed over local data sets $\mathcal{D}^k_i$ sampled from the same target distribution $P_i$. The variances are computed as a function of the partitioned mini-batch sizes. Consequently, the size of the mini-batches shall be chosen carefully to successfully capture clients of similar data distributions during training. We discuss the suitable choice of the mini-batch sizes to compute the variances in section \ref{variance}. Once all the necessary quantities are computed, they are uploaded to the PS, which proceeds to calculate the mixture coefficients and initiates the federated training using the custom aggregation scheme given by (\ref{eq3}). An illustration of our proposal is found in Algorithm \ref{Algo2}.

Note that the proposed heuristic embodies the intuition provided by Theorem \ref{Th1}. In fact, in the case of homogeneous users, it falls back to the standard FedAvg aggregation rule, while if node $i$ has an infinite amount of data it degenerates to the local learning rule which is optimal in that case.

\begin{algorithm}[ht]
\algsetup{linenosize=\small}
\SetAlgoLined
\DontPrintSemicolon
\SetKwInOut{Input}{Input}
\SetKwInOut{Output}{Output}
  \caption{User-centric Federated Learning}
  \Input{number of clients $m$, local mini-batch size $B$, number of epochs $E$ and learning rate $\eta$}
       
       \textnormal{PS} broadcasts $\theta^0$ to the users\;
        \ForEach{\textnormal{user} $k$} { \textnormal{Compute} $\nabla\ell(\theta^0,\mathcal{D}_k)$   
        \;$\text{Compute $\sigma^2_k$  as in } \text{(\ref{eq9})}$\;
        \text{Transmit }$\{\nabla\ell(\theta^0,\mathcal{D}_k), \sigma^2_k\}$ \text{to PS}    }
        \textnormal{PS computes} $w_{i,j}$ \text{as in (\ref{eq7})}  \;
        \For {$t = 0,\dots,T$} {
          $\textnormal{PS } \text{unicasts } \theta_k^t \text{ to each node } k$ \; 
                          \ForEach{node $k$ } { $\theta^{t+1}_k \gets \textbf{ClientUpdate}(\theta^t_k,\mathcal{D}_k)$ \; $\text{return } \theta^{t+1}_k \text{ to } \textit{PS}$     }
          $\text{PS computes } \theta_k^{t+1} \gets \sum_{j=1}^m w_{k,j}\theta_j^{t+1}$}
        \;
        $\textbf{PROCEDURE: }$$\textnormal{ClientUpdate($\theta^t_k,\mathcal{D}_k$):}$  \\
        $\mathcal{B}$ $\gets$ $\textnormal{Split}$ $\mathcal{D}_k$ $\textnormal{into batches of size $B$}$
        \;$\theta_k \gets \theta_k^{t}$
        \;\For{$t=0,\dots,E$}
            { \ForEach{\textnormal{batch $b$} $\in \mathcal{B}$}
                {   $\theta_k$ $\gets$ $\theta_k-\eta \nabla \ell(\theta_k,b)$
                }}
        $\textbf{return}$ $\theta_k$
\label{Algo2}
\end{algorithm}
\subsection{Reducing the Communication Load} \label{Clustering}
A full-fledged personalization employing the user-centric aggregation rule (\ref{eq3}) would introduce an $m$-fold increase in communication load during the downlink phase as the original broadcast transmission is replaced by unicast ones. Although from a learning perspective the user-centric learning scheme is beneficial, it is also possible to consider overall system performance from a learning-communication trade-off point of view. The intuition is that, for small discrepancies between the user data distributions, the same model transfer positively to statistically similar devices. 
To strike a suitable trade-off between learning accuracy and communication overhead we hereby propose to adaptively limit the number of personalized downlink streams. In particular, for a number of personalized models $m_t$, we run a $k$-means clustering scheme over the set of collaboration vectors $\{\vec{w}_i\}_{i=1}^m$ and we select the centroids $\{\vec{c}_i\}_{i=1}^{m_t}$ to implement the $m_t$ personalized streams. Formally, given $m_t$ and the user-specific weights $\{\vec{w}_{i}\}^m_{i=1}$, the objective is to find $m_t<m$ clusters $\mathcal{C}_1,\dots,\mathcal{C}_{m_t}$ such that\begin{equation}
\sum_{n=1}^{m_t} \sum_{\vec{w}_i \in \mathcal{C}_n}\norm{\vec{w}_i - \vec{c}_n}
\end{equation}
is minimized, where $\vec{c}_n$ is the centroid of cluster $\mathcal{C}_{n}$. We then proceed to replace the unicast transmission with group broadcast ones, in which all users belonging to the same cluster $i$ receive the same personalized model associated with the centroid $\vec{c}_i$. Choosing the right value for the number of personalized streams is critical to save communication bandwidth but at the same time obtain satisfactory personalization capabilities. In the following, we experimentally show that clustering quality indicators such as the Silhouette score can be used to guide the search for a suitable number of clusters $m_t$.

\subsection{Choosing the Number of Personalized Streams}

\begin{algorithm}[ht]
\algsetup{linenosize=\small}
\SetAlgoLined
\DontPrintSemicolon
\SetKwInOut{Input}{Input}
\SetKwInOut{Output}{Output}
  \caption{Silhouette based scoring}
  \Input{ Collaboration vectors $\{\vec{w}_{i}\}^m_{i=1}$ from Algorithm \ref{Algo2} and a trade-off function $c(k,s_k)$.} 
  
  \Output{ Number of clusters $m_t$}
 
  \For{$K = 1,2,\dots,m$}
  {
  $\mathcal{C}_k\leftarrow$ $K$-means clustering of $\{\vec{w}_{i}\}^m_{i=1}$\\
  $s_k\leftarrow$   the silhouette score of $s(\mathcal{C}_k)$
  }
  \textbf{return }$\argmax_{k=1,\dots,m} c(k,s_k)$
\label{alg3}
\end{algorithm}

\label{abc}
Choosing an insufficient number of personalized streams can yield unsatisfactory performance, while concurrently learning many models can prohibitively increase the communication load of personalized federated learning. Therefore, properly tuning this free parameter is essential to obtain a well-performing but still practical algorithm. Being agnostic w.r.t. the underlying data generating distributions at the devices, it does not exist a universal number of personalized streams that fits all problems. However, we now illustrate that the silhouette coefficient, a quality measure of the clustering, provides a rule of thumb to choose the number of personalized streams. In order to compute the silhouette score of a clustering $\mathcal{C}_1,\dots,\mathcal{C}_{m_t}$ of the clustering we define the intra-cluster similarity of the collaboration vector $\vec{w}_i \in \mathcal{C}_k$ as
\[
a(\vec{w}_i)=\frac{1}{|\mathcal{C}_j|-1}\sum_{\vec{w}_j\in \mathcal{C}_k,\vec{w}_j\neq\vec{w}_i}\norm{\vec{w}_j-\vec{w}_i}
\]
and the smallest mean distance between the collaboration vector $\vec{w}_i \in \mathcal{C}_k$ and the closest cluster
\[
b(\vec{w}_i)=\min_{\mathcal{C}_j\neq \mathcal{C}_k}\frac{1}{|\mathcal{C}_j|}\sum_{\vec{w}_j\in \mathcal{C}_j}\norm{\vec{w}_j-\vec{w}_i}.
\]
The average silhouette score $s$ is then defined as
\[
s(\mathcal{C})=\frac{1}{m}\sum_{i=1}^m \frac{b(i)-a(i)}{\max\{a(i),b(i)\}}
\]
and it is a number in the range $\left[-1, 1\right]$, directly proportional to the quality of the clustering. In turn, a good clustering of the collaboration vectors $\{\vec{w}_i\}^m_{i=1}$ implies that users belonging to the same clusters are similar and that the centroid $\vec{c}_j$ is a good approximation of the collaboration coefficient of users in $\mathcal{C}_j$. Consequently, whenever the silhouette score is large, the loss in terms of personalization performance resulting from the reduced number of aggregation rules compared to the full-fledged personalization system is modest. For this reason, the silhouette score provides a proxy to the inference performance and at the same time, it allows to trade-off communication load and personalization capabilities in a principled way.
In Algorithm \ref{alg3} we provide the pseudocode of the procedure that autonomously chooses the optimal number of personalized streams $m_t$ based on a communication-personalization trade-off function $c(k,s):\mathbb{N}\times[-1,1]\to \mathbb{R}^+$ scoring the utility of pairs of the systems based on the number of user-centric rules and the resulting silhouette scores. The function $c(k,s)$ is a system dependent function typically decreasing in $k$ and increasing in $s_k$. 

\begin{figure*}[ht!]
\centering

    \begin{subfigure}[t]{0.329\textwidth}
         \centering
         \includegraphics[width=\textwidth]{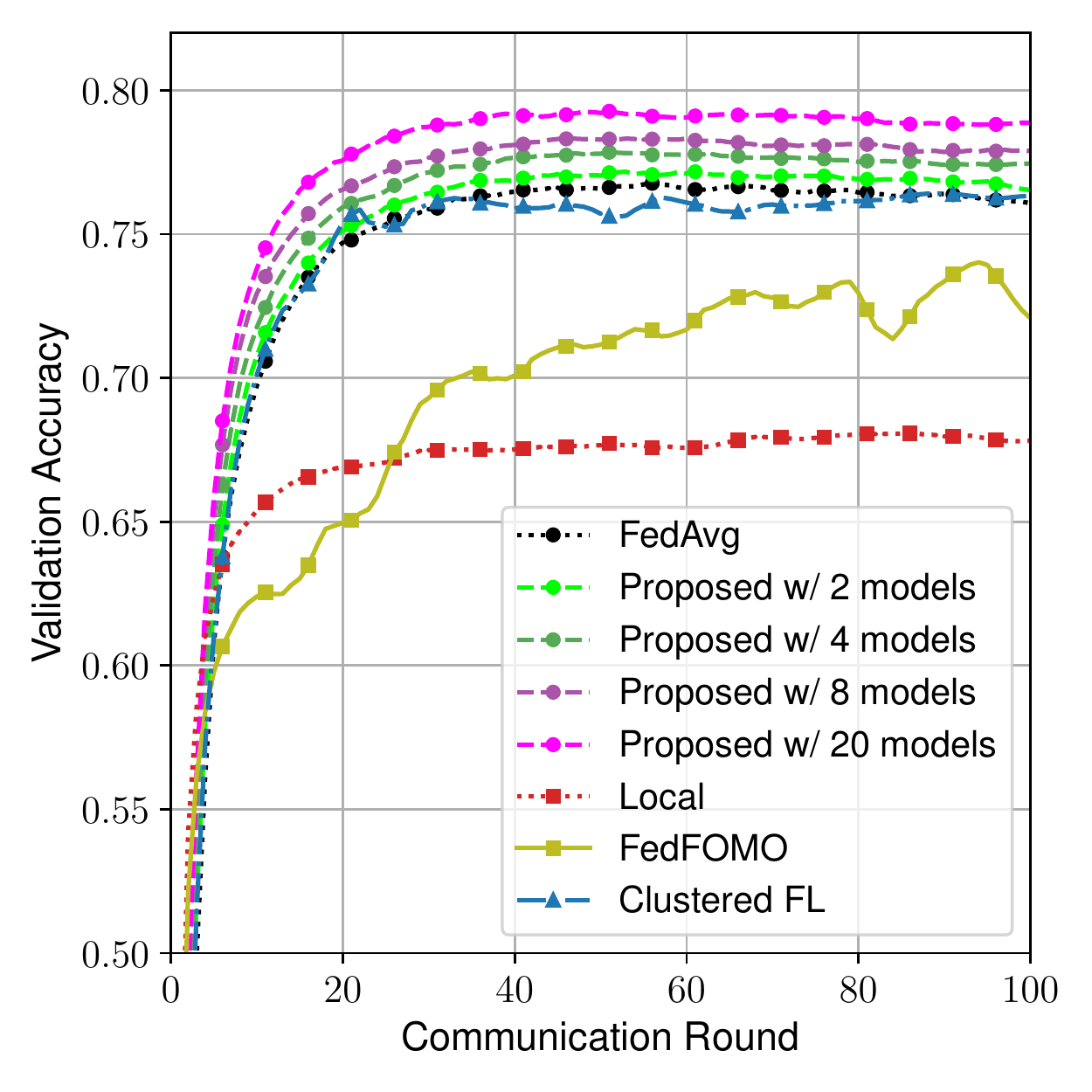}
         \centering
         \caption{EMNIST + label shift}
         \label{fig:EMNIST_label}
     \end{subfigure}
      \begin{subfigure}[t]{0.329\textwidth}
         \centering
          \includegraphics[width=\textwidth]{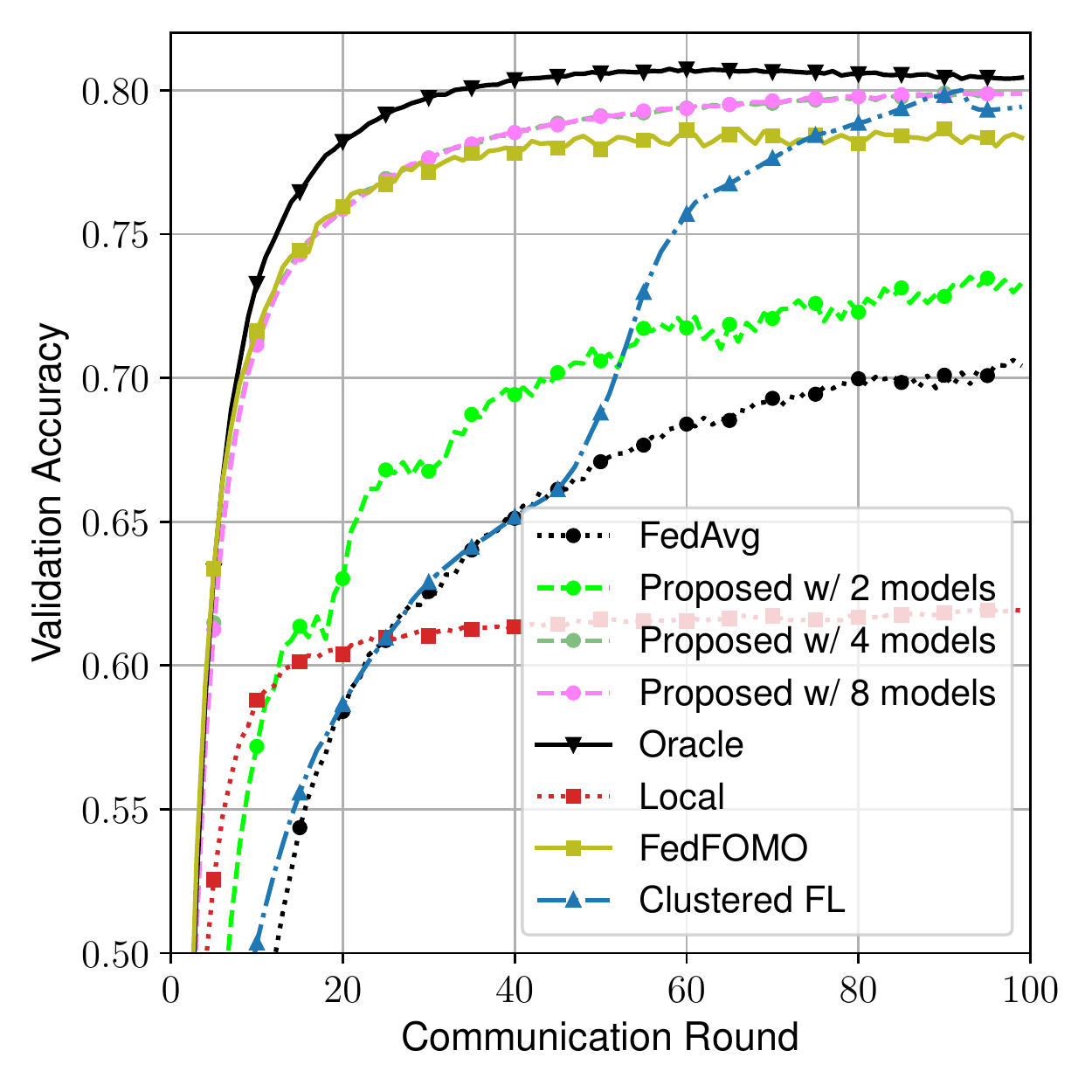}
         \caption{EMNIST + label and covariate shift}
         \label{fig:EMNIST_cov}
     \end{subfigure}
     \begin{subfigure}[t]{0.329\textwidth}
         \centering
          \includegraphics[width=\textwidth]{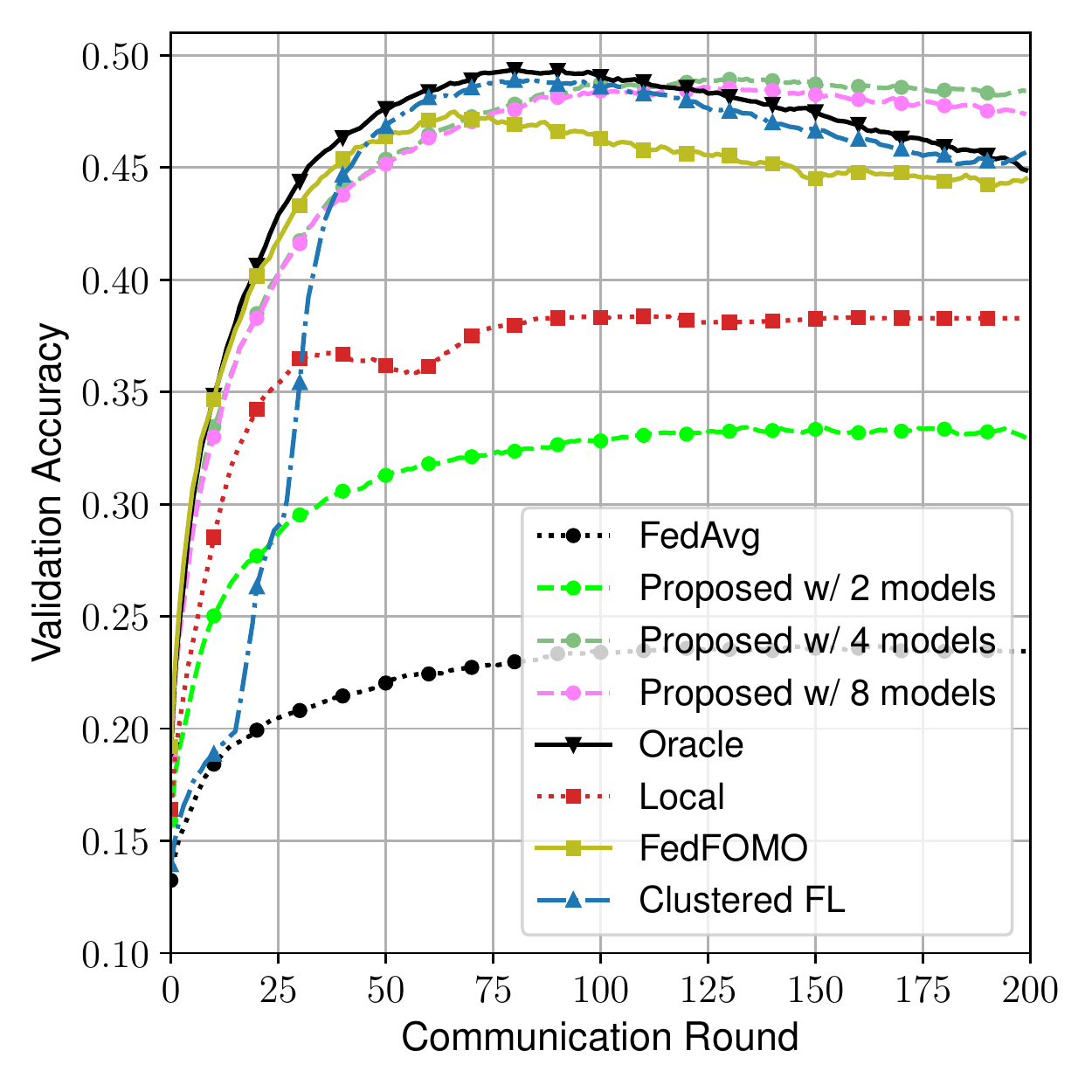}
         \caption{CIFAR10 + concept shift}
         \label{fig:CIFAR}
     \end{subfigure}
     \caption{Average Validation Accuracy across the three different experiements}
     \label{acc}
\end{figure*}

\begin{table*}
\caption{Average test accuracy of the different algorithms across the three proposed scenarios.}
\centering
\small
\begin{tabular}{ @{}lccc @{} }
\toprule
  \multicolumn{1}{c}{Algorithm}    & \multicolumn{3}{c}{Scenario}              \\\cmidrule(lr){2-4}
 \centering  & \thead{\small EMNIST ($m$ = 20)\\ \small label shift}     & \thead{\small EMNIST ($m$ = 100)\\\small covariate \& label shift} & \thead{\small CIFAR10 ($m$ = 20)\\\small concept shift} \\
 \midrule
 Proposed  $k=m$   &\textbf{79.4} ($\pm$ 4.2) & \textbf{77.9} ($\pm$ 2.7) &  \textbf{47.7} ($\pm$ 2.2)\\
 Proposed $k=4$     &77.8 ($\pm$ 3.9)& \textbf{79.7} ($\pm$ 2.5)&  \textbf{49.1} ($\pm$ 1.4)\\
 SCAFFOLD \cite{DBLP:journals/corr/abs-1910-06378} &77.2 ($\pm$ 4.0) & 72.5 ($\pm$ 2.2)& 17.5 ($\pm$ 1.8)\\
 Ditto \cite{DBLP:journals/corr/abs-2012-04221}    &78.3 ($\pm$ 3.9) & 74.1 ($\pm$ 2.3)&  44.1 ($\pm$ 1.4)\\
 pFedMe \cite{DBLP:journals/corr/abs-2006-08848}&   77.6 ($\pm$ 4.1)  & 75.2($\pm$ 4.4)&46.6 ($\pm$ 1.5)\\
 Fedprox \cite{DBLP:journals/corr/abs-1812-06127}& \textbf{79.6} ($\pm$ 4.8)  & 72.4 ($\pm$ 2.4)   &22.3 ($\pm$ 2.2) \\
 Local& 68.2 ($\pm$ 5.3)  & 62.8 ($\pm$ 3.3)  &38.3 ($\pm$ 1.2)\\
 FedAvg \cite{mcmahan2017communication}& 76.7 ($\pm$ 4.0)  & 70.5 ($\pm$ 2.2)&24.2 ($\pm$ 2.6)\\
  Oracle \emph{(Upper bound)}&   -  & \textcolor{red}{80.7} ($\pm$ 1.8) &\textcolor{red}{49.5} ($\pm$ 1.2)\\
\bottomrule
\end{tabular}%

\label{table1:acc}
\end{table*}
\begin{table*}[ht!]
\centering
\small
\caption{ \centering Worst user performance averaged over 5 experiments in the three simulation scenarios}
\label{table:worst}
\begin{tabular}{@{} l  c  c  c  c  c  c  c  @{}}
\toprule 
\multicolumn{1}{c}{Scenario}  & \multicolumn{5}{c}{Algorithm}              \\\cmidrule(lr){2-8}
& Ditto \cite{DBLP:journals/corr/abs-2012-04221} & FedAvg \cite{mcmahan2017communication} & Oracle & CFL \cite{sattler2020clustered} & FedFOMO \cite{zhang2020personalized} & pFedMe \cite{DBLP:journals/corr/abs-2006-08848} & Proposed\\
\midrule
$\text{EMNIST ($m$ = 20) label shift }$ & 72.2 & 68.9 & - & 70.3 & 70.0 & 71.5 & \textbf{73.2} $(k=20)$\\

$\text{EMNIST ($m$ = 100) covariate \& label shift}$ & 70.7 & 67.5 & 77.4 & 76.1 &73.6 &70.9& \textbf{76.4} $(k=4)$\\

$\text{CIFAR10  ($m$ = 20) concept shift}$ & 43.2 & 19.6 & 49.1 & 48.6 & 45.5 & 45.3 & \textbf{48.8} $(k=4)$\\
\bottomrule
\end{tabular}

\end{table*}

\section{Experiments}
\label{section5}
We now provide a series of experiments to showcase the personalization capabilities and communication efficiency of the proposed algorithm.
\subsection{Set-up}
\label{setup}
In our simulation we consider a handwritten character/digit recognition task using the EMNIST data set \cite{cohen2017emnist}  and an image classification task using the CIFAR-10 data set \cite{krizhevsky2009learning}.
Data heterogeneity is induced by splitting and transforming the data set differently across the group of devices. In particular, we analyze three different scenarios:
\begin{itemize}
    \item\textbf{Character/digit recognition with user-dependent label shift} in which 10k EMNIST data points are split across 20 users according to their labels. The label distribution follows a Dirichlet distribution with parameter $\alpha =$ 0.4, as in \cite{mixture2021s,wang2020tackling}.
    
    \item\textbf{Character/digit recognition with user-dependent label shift and covariate shift} in which 100k samples from the EMNIST data set are partitioned across 100 users each with a different label distribution ($\alpha = 8$), as in the previous scenario. Additionally, users are clustered in 4 groups $\mathcal{G} = \{\mathcal{G}_1,\mathcal{G}_2,\mathcal{G}_3,\mathcal{G}_4\}$, and at each group images are rotated by $\{0^{\circ},90^{\circ},180^{\circ},270^{\circ}\}$ respectively. In particular, heterogeneity is imposed such that $p_i\left(x | y\right) \ne p_j(x | y), \,\forall \, i\in \mathcal{G}_k,j\in\mathcal{G}_{k'}, k\ne k' ,\, \forall (x,y) \in \mathcal{X}\times\mathcal{Y}$.
    
     \item\textbf{Image classification with group dependent concept shift} in which the CIFAR-10 data set is distributed across 20 users which are grouped in 4 clusters, for each group we apply a different random label permutation. More specifically, given an image $x \in \mathcal{X}$ and the labelling functions $f_i,f_j: \mathcal{X} \rightarrow \mathcal{Y}$, then $f_i(x) \ne f_j(x), \forall i\in \mathcal{G}_k\,$,$\,j\in\mathcal{G}_{k'}, k\ne k'$.
\end{itemize}

For each scenario, we aim at solving the task at hand by leveraging the distributed and heterogeneous data sets. We compare our algorithm against two sets of baseline algorithms. The first set includes algorithms that achieve personalization by resulting multiple personalized models. Those include CFL \cite{sattler2020clustered}, FedFomo \cite{zhang2020personalized}, pFedMe \cite{DBLP:journals/corr/abs-2006-08848} and Ditto \cite{DBLP:journals/corr/abs-2012-04221}. The second set of baselines include algorithms that yield a single Federated model such as Fedprox \footnote{\textnormal{The penalizationn hyperparameters $\mu\textnormal{ and } \lambda = \{0.1,0.5,1\}$ were used in the simulations of Fedprox and Ditto, then, the best results were reported.}}\cite{DBLP:journals/corr/abs-1812-06127}, SCAFFOLD \cite{DBLP:journals/corr/abs-1910-06378}. FedAvg \cite{mcmahan2017communication}, and Local training algorithms are also included for reference. All algorithms are trained using LeNet-5  \cite{lecun1998gradient} convolutional neural network. In all scenarios and for all algorithms\footnote{\textnormal{\textnormal{Exception:} The hyperparameters $ \eta_{global}=\eta_{local} = 0.01$, $S = 15$, $E=1$ and batchsize$ = 20$ were used for pFedMe, and $\eta=0.01$, $E = 5$ for SCAFFOLD}}, we use stochastic gradient descent optimizer with fixed learning rate $\eta=0.1$, momentum $\beta=0.9$, and the number of epochs $E=1$.


\subsection{Personalization Performance}
\label{sec:per_perf}
We now report the average accuracy over 5 trials attained by the different approaches. We also study the personalization performance of our algorithm when we restrain the overall number of personalized streams, namely the number of personalized models that are concurrently learned. \subsubsection{Multi-Model Baseline Algorithms}
In Fig.\ref{acc} and Table \ref{table1:acc}, we report the average validation accuracy of the baseline algorithms that yield multiple personalized models, alongside FedAvg, Fedprox, SCAFFOLD and local training. In the EMNIST label shift scenario (Fig.\ref{fig:EMNIST_label}), we first notice that harvesting intelligence from the data sets of other users amounts to a large performance gain compared to the localized learning strategy. This indicates that data heterogeneity is moderate and collaboration is fruitful.  Nonetheless, personalization can still provide gains compared to FedAvg. Our solution yields a validation accuracy which is increasing in the number of personalized streams. Allowing maximum personalization, namely a different model for each user, we obtain a 3\% gain in the average accuracy compared to FedAvg. CFL is not able to transfer intelligence among different groups of users and attains performance similar to the FedAvg. This behaviour showcases the importance of soft clustering compared to the hard one for the task at hand. We find that FedFOMO, despite excelling in case of strong statistical heterogeneity, fails to harvest intelligence in the label shift scenario. In Fig.\ref{fig:EMNIST_cov} we report the personalization performance for the second scenario. In this case, we also consider the oracle baseline, which corresponds to running 4 different FedAvg instances, one for each cluster of users, as if the 4 groups of users were known beforehand.  Different from the previous scenario, the additional shift in the covariate space renders personalization necessary to attain satisfactory performance. The oracle training largely outperforms FedAvg. Furthermore, as expected, our algorithm matches the oracle final performance when the number of personalized streams is 4 or more. Also, CLF and FedFOMO can correctly identify the 4 clusters. However, the former exhibits slower convergence due to the hierarchical clustering over time while the latter plateaus to a lower average accuracy level. We turn now to the more challenging CIFAR-10 image classification task. In Fig.\ref{fig:CIFAR} we report the average accuracy of the proposed solution for a varying number of personalized streams, the baselines and the oracle solution. As expected, the label permutation renders collaboration extremely detrimental as the different learning tasks are conflicting. As a result, local learning provides better accuracy than FedAvg. On the other hand, personalization can still leverage data among clusters and provide gains also in this case. Our algorithm matches the oracle performance for a suitable number of personalized streams. This scenario is particularly suitable for hard clustering, which isolates conflicting data distributions. As a result, CFL matches the proposed solution. FedFOMO promptly detects clusters and therefore quickly converges, but it attains lower average accuracy compared to the proposed solution. On the other hand, Ditto and pFedMe  perform relatively better than the aforementioned two approaches, given their personalization capabilities. However, they fall short while leveraging collaboration among users towards training the global model only, and disregarding the potential generalization gain that could be achieved by enabling collaboration among statistically similar users towards refining their local personalized models. 
\subsubsection{Single-Model Baseline Algorithms}

Despite that all algorithms that yield a single model (i.e. Fedprox and SCAFFOLD) excel  in the label shift setting (Table \ref{table1:acc}), our proposed algorithm stands out in the two other scenarios. This stems from their inadequacy in addressing the conflicting nature of the available target tasks via a single global model in the other two scenarios. 

\begin{figure*}[h]
\centering
    \begin{subfigure}[t]{0.49\textwidth}
         \centering
         \includegraphics[width=\textwidth]{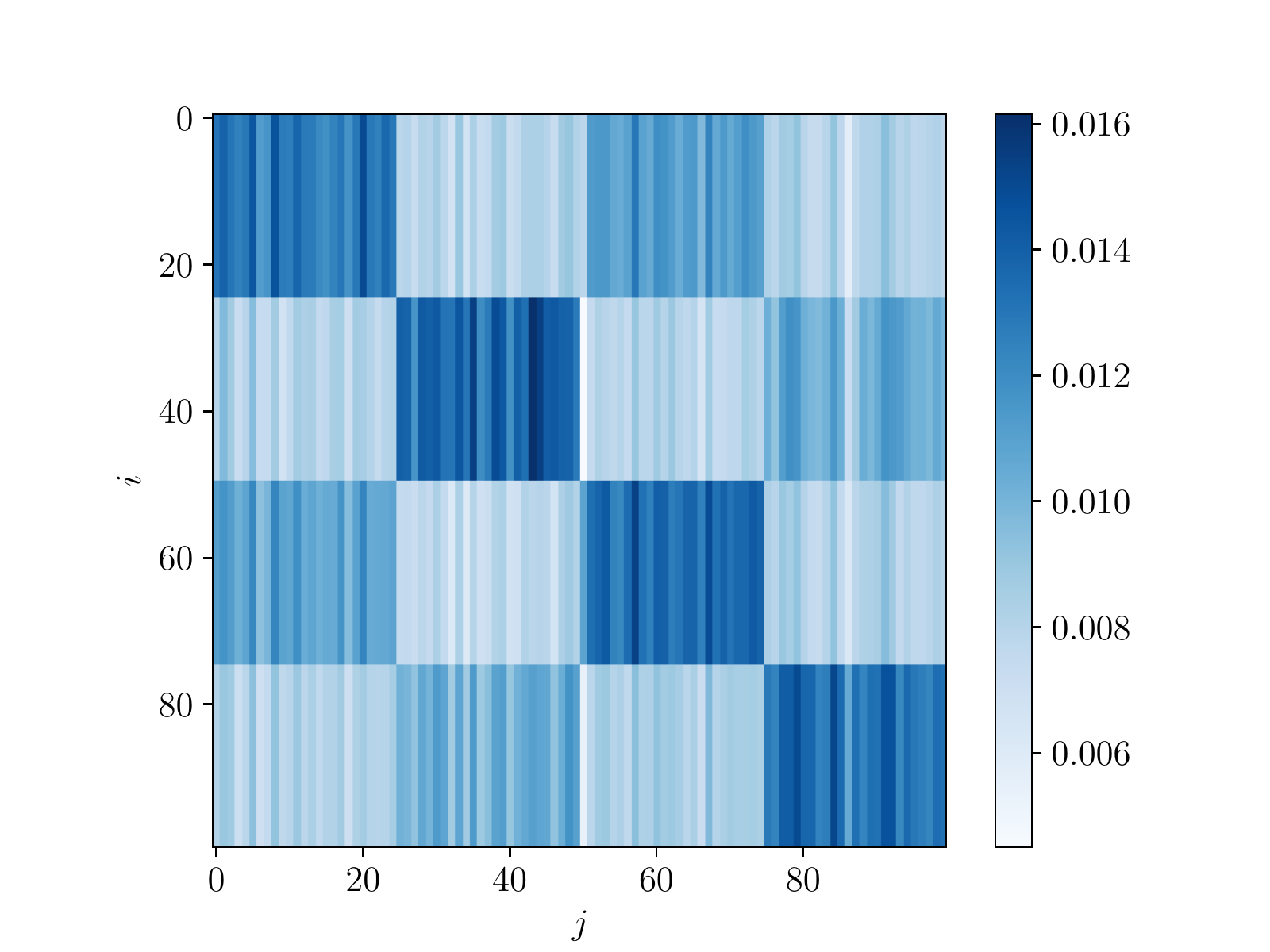}
         \centering
         \caption{\centering EMNIST label \& covariate shift (100 clients).}
         \label{fig:clusters1}
     \end{subfigure}
      \begin{subfigure}[t]{0.49\textwidth}
         \centering
         \includegraphics[width=\textwidth]{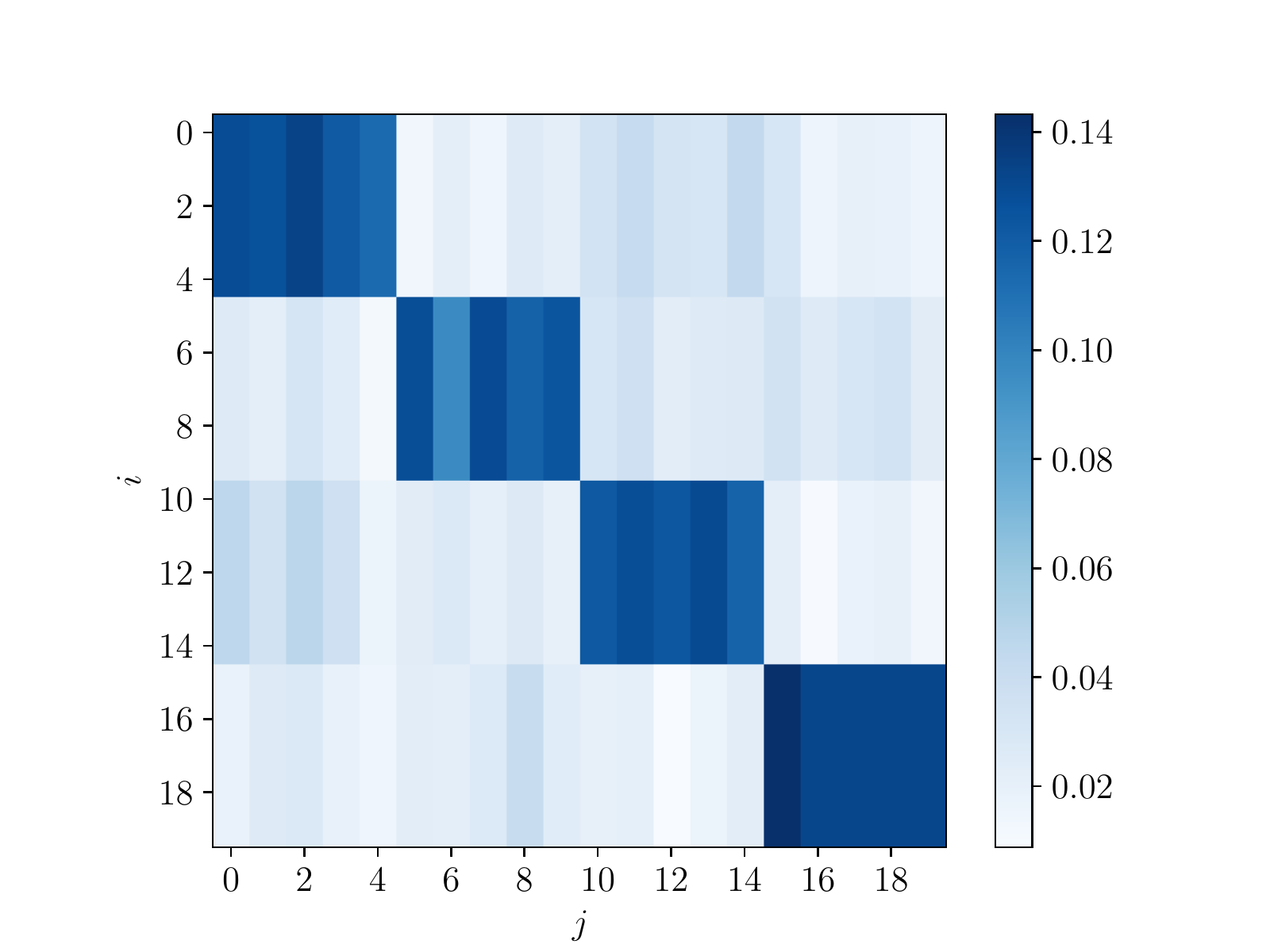}
         \caption{\centering CIFAR10 concept shift (20 clients).}
         \label{fig:clusters2}
     \end{subfigure}
     \caption{\centering Clusters formed by our proposed algorithm in the EMNIST label and covariate shift, and CIFAR10 concept shift scenarios. Each 2D point denotes $w_{i,j}$ : A dark blue point $w_{i,j}$ conveys a relatively large collaboration between user $i$ and $j$.}
     \label{fig:clusters}
\end{figure*}

\subsubsection{Average Worst Performance}
The performance reported so far is averaged over users and therefore fails to capture the existence of outliers performing worse than average. To assess the fairness of the training procedure, in Table \ref{table:worst} we report the worst user performance in the federated system across the different algorithms. The proposed approach produces models with the highest worst case in all three scenarios.
\subsubsection{Inter-Cluster Collaboration}
We illustrate the clustering performance of our proposed solution in the EMNIST co-variate shift and the CIFAR10 concept shift scenarios (Experiments two and three) with four clusters each in Fig. \ref{fig:clusters}. Interestingly, we notice that in the EMNIST covariate shift experiment, our clustering algorithm can detect similarities among the different groups of users, leveraging inter-cluster collaboration among them, unlike hard clustering algorithms \cite{sattler2020clustered}. This stems from the fact that some digits and letters features are invariant to the 180$^{\circ}$ rotation applied  (e.g letters $X,Z,O,N,\textnormal{etc}\,...$ and the digits $\{0,1,8\}$).


\subsection{Silhouette Score}
\begin{figure*}[]
\centering
    \begin{subfigure}[t]{0.31\textwidth}
         \centering
         \includegraphics[width=\textwidth]{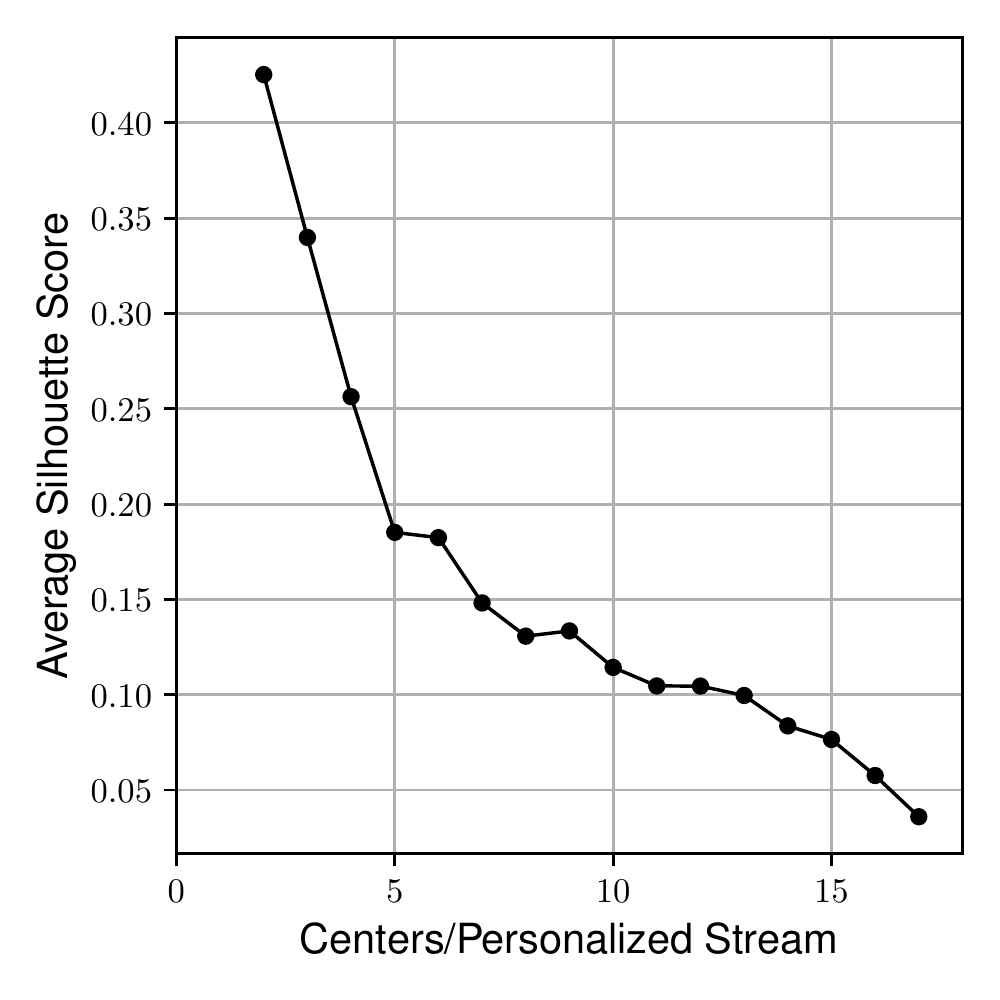}
         \centering
         \caption{EMNIST label shift.}
         \label{fig:EMNIST_labelsol}
     \end{subfigure}
      \begin{subfigure}[t]{0.31\textwidth}
         \centering
         \includegraphics[width=\textwidth]{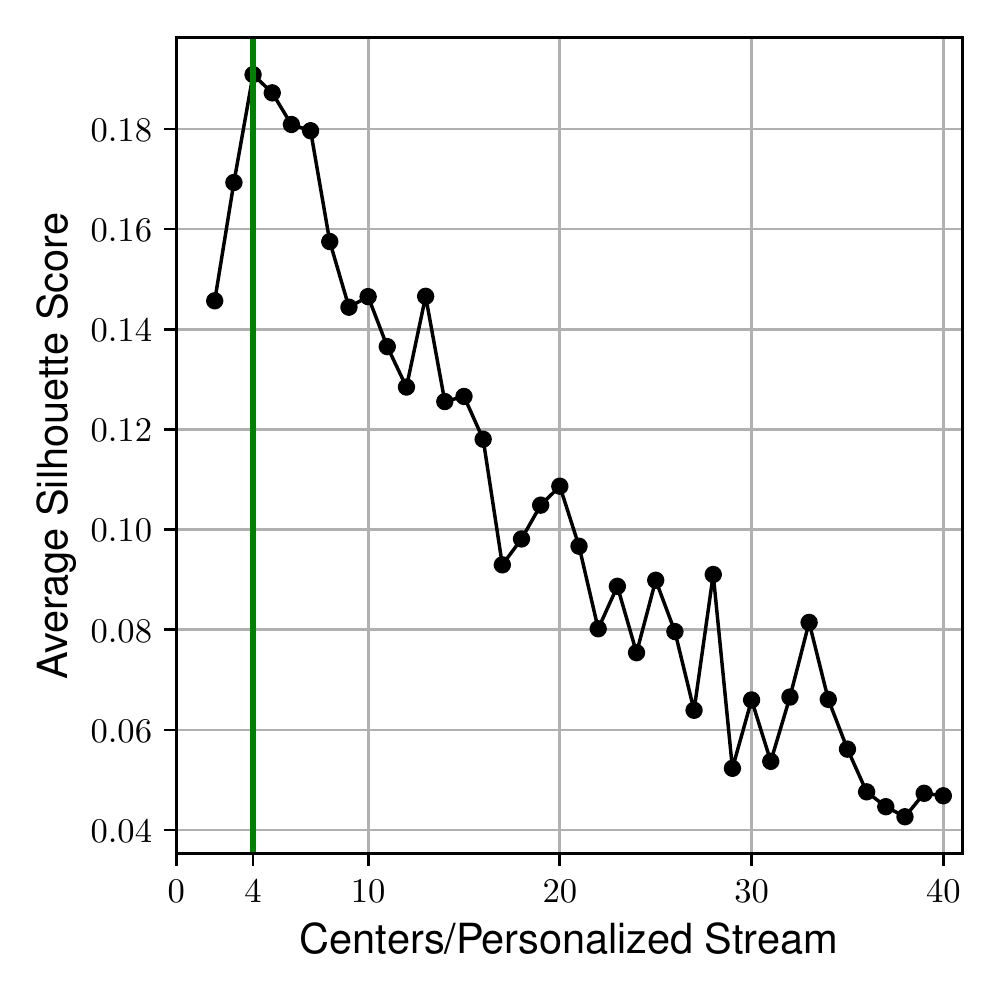}
         \caption{EMNIST label \& covariate shift.}
         \label{fig:EMNIST_covsol}
     \end{subfigure}
     \begin{subfigure}[t]{0.31\textwidth}
         \centering
          \includegraphics[width=\textwidth]{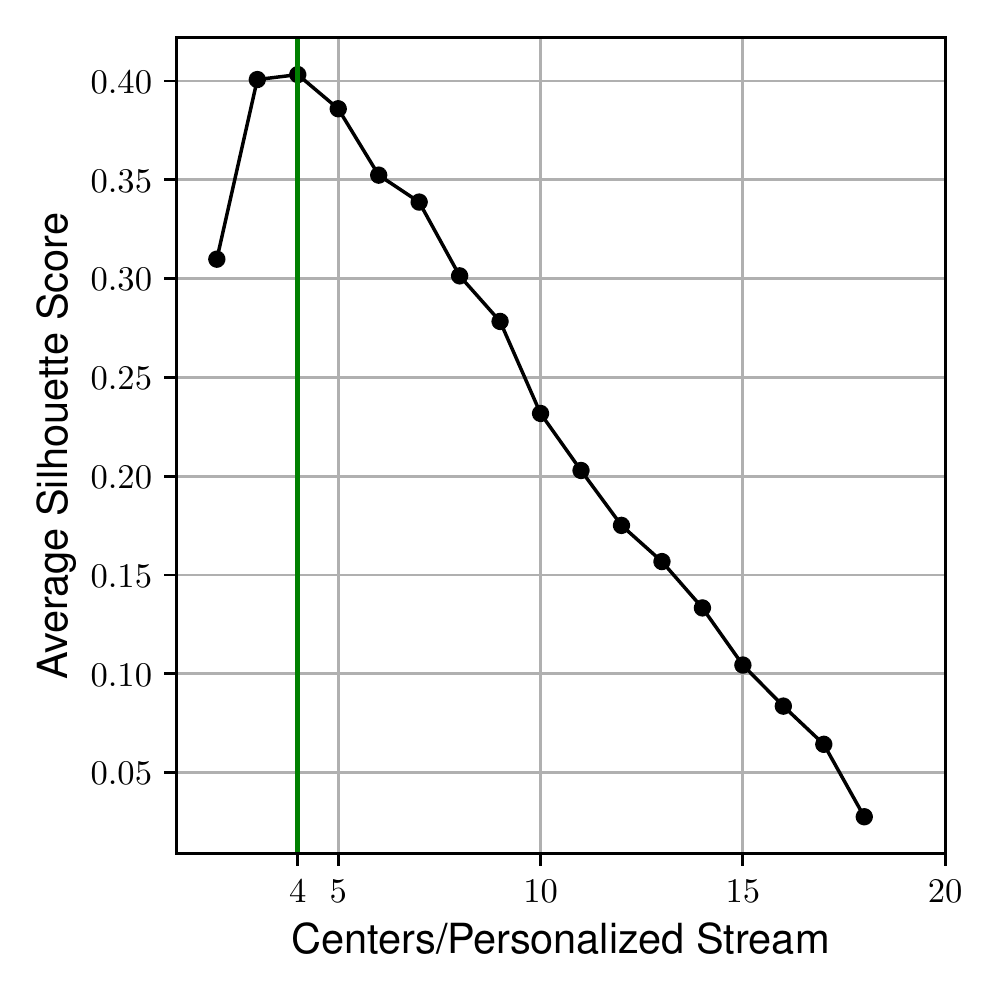}
         \caption{CIFAR10 concept shift.}
         \label{fig:CIFARsol}
     \end{subfigure}
     \caption{\centering Average silhouette scores of the $k$-means clustering in the three scenarios. In the last two scenarios, in which user inherently belongs to 4 different cluster, the scores indicates the necessity of at least 4 personalized streams.}
     \label{fig:Silhouette}
\end{figure*}

In Fig. \ref{fig:Silhouette} we plot the average silhouette score obtained by the $k$-means algorithm when clustering the federated users based on the procedure proposed in Sec. \ref{Clustering}. In the labels shift scenario, for which we have seen that a universal model performs almost as well as the personalized ones, the silhouette scores monotonically decrease with $k$. In fact, in this simulation setting, a natural cluster-like structure among clients' tasks does not exist. On the other hand, in the covariate shift and the concept shift scenarios, the silhouette score peaks around $k=4$. In Sec. \ref{sec:per_perf} this has shown to be the minimum number of personalized models necessary to obtain satisfactory personalization performance in the system. This behaviour of the silhouette score is expected and desired, in this case, the number of clusters matches exactly the number of underlying different tasks among the participants in FL that was induced by the rotation of the covariates and the permutation of the labels. We then conclude that the silhouette score provides meaningful information to tune the number of user-centric aggregation rules before training.
\label{variance}

\subsection{Communication Efficiency}

\begin{figure*}[]
\centering
    \begin{subfigure}[t]{0.31\textwidth}
         \centering
         \includegraphics[width=\textwidth]{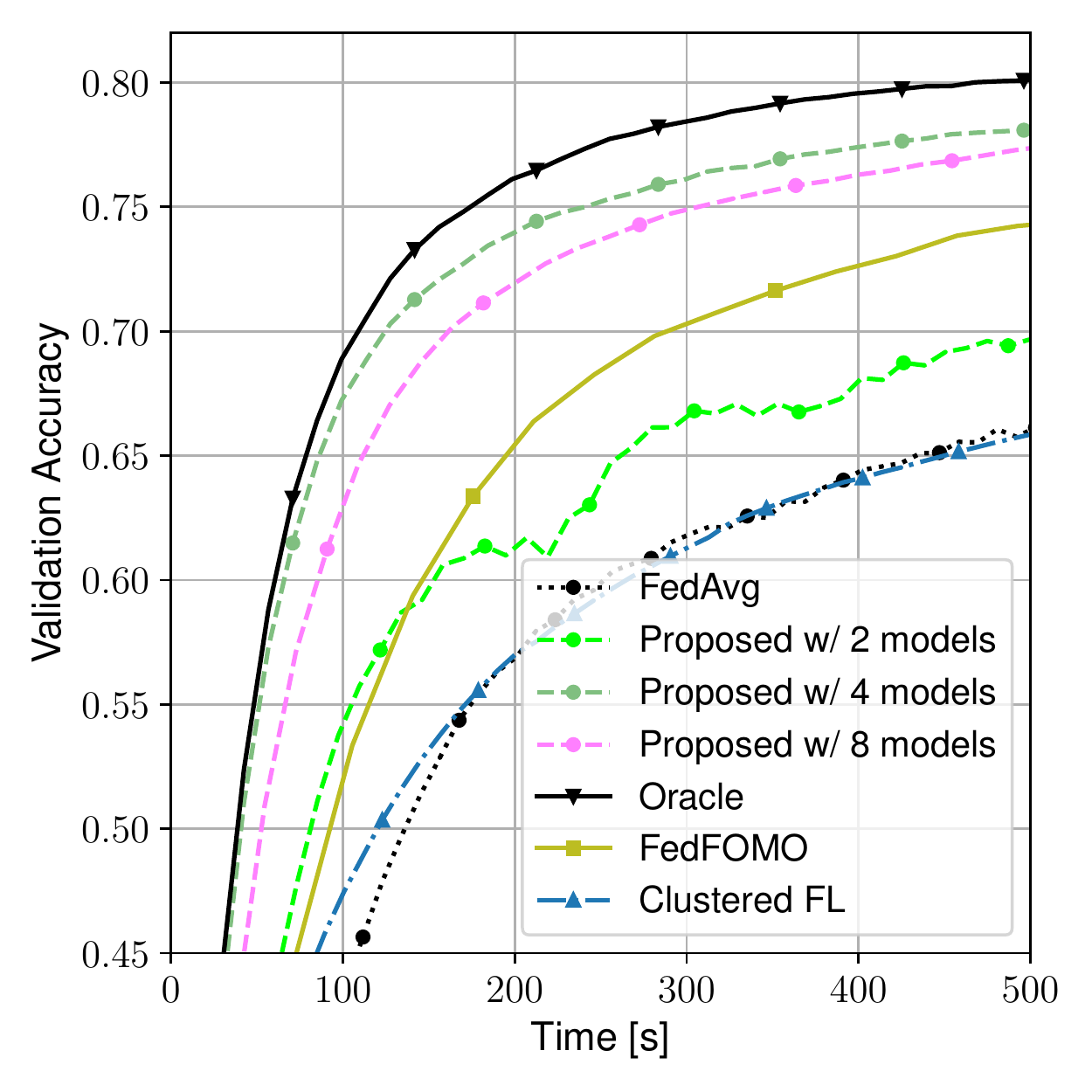}
         \centering
         \caption{$\rho=4, T_{min}=T_{dl}=\frac{1}{\mu}$}
         \label{fig:rho4}
     \end{subfigure}
      \begin{subfigure}[t]{0.31\textwidth}
         \centering
         \includegraphics[width=\textwidth]{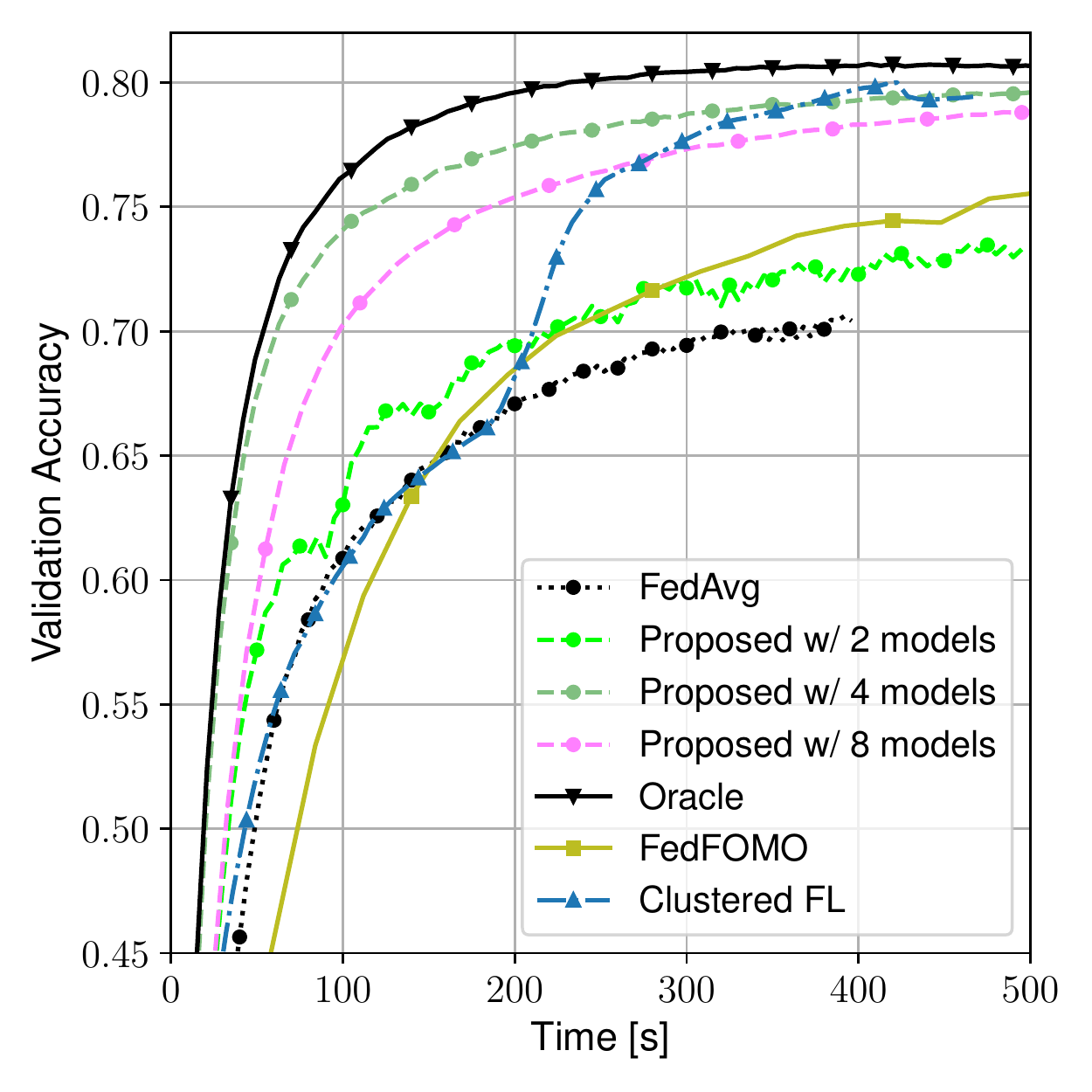}
         \caption{$\rho=2, T_{min}=T_{dl},\frac{1}{\mu}=0$}
         \label{fig:rho2}
     \end{subfigure}
     \begin{subfigure}[t]{0.31\textwidth}
         \centering
          \includegraphics[width=\textwidth]{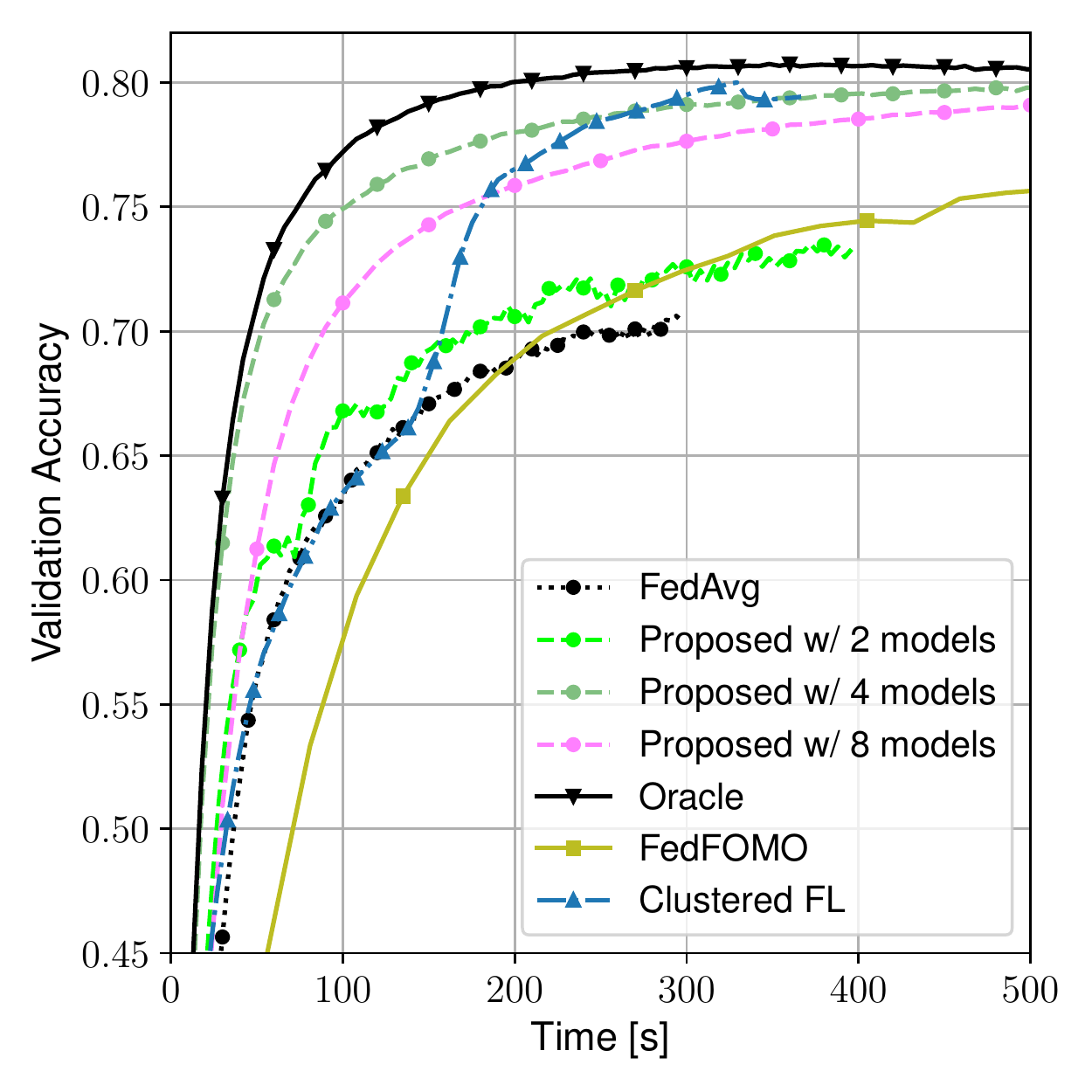}
         \caption{$\rho=1, T_{min}=T_{dl},\frac{1}{\mu}=0$}
         \label{fig:rho3}
     \end{subfigure}
     
     \caption{Evolution of the average validation accuracy against time normalized w.r.t. $T_{dl}$ for the three different systems.}
     \label{fig:comm_eff}
\end{figure*}
Personalization comes at the cost of increased communication load in the downlink transmission from the PS to the federated user. To compare the algorithm convergence time, we parametrize the distributed system using two parameters. We define by $\rho=\frac{T_{ul}}{T_{dl}}$ the ratio between model transmission time in uplink (UL) and downlink (DL). Typical values of $\rho$ in wireless communication systems are in the $[2,4]$ range because of the larger transmitting power of the base station compared to the edge devices. Furthermore, to account for unreliable computing devices, we model the random computing time $T_i$ at each user $i$ by a shifted exponential r.v.  with a cumulative distribution function
\[
P[T_i>t]=1-\mathbbm{1}(t\geq T_{min})\left[1-e^{-\mu(t-T_{min})}\right]
\]
where $T_{min}$ represents the minimum possible computing time and $1/\mu$ is the average additional delay due to random computation impairments. Therefore, for a population of $m$ devices, we then have 
\[
T_{comp}=\mathbb{E}\left[\max\{T_1,\dots,T_m\}\right]=T_{min}+\frac{H_m}{\mu}
\]
where $H_m$ is the $m$-th harmonic number.
To study the communication efficiency we consider the simulation scenario with the EMNIST data set with label and covariate shift.  
In Fig. \ref{fig:comm_eff} we report the time evolution of the validation accuracy in 3 different systems. A wireless systems with slow UL $\rho=4$ and unreliable nodes $T_{min}=T_{dl}=\frac{1}{\mu}$, a wireless system with fast uplink $\rho=2$ and reliable nodes $T_{min}=T_{dl}$, $\frac{1}{\mu}=0$ and a wired system $\rho=1$ (symmetric UL and DL) with reliable nodes $T_{min}=T_{dl}$, $\frac{1}{\mu}=0$. The increased DL cost is negligible for wireless systems with strongly asymmetric UL/DL rates and in these cases, the proposed approach largely outperforms the baselines. In the case of more balanced UL and DL transmission times $\rho=[1,2]$ and reliable nodes, it becomes instead necessary to properly choose the number of personalized streams to render the solution practical. Nonetheless, the proposed approach remains the best even in this case for $k=4$. Note that FedFOMO incurs a large communication cost as personalized aggregation is performed on the client side.

\begin{figure*}[h]
\centering
    \begin{subfigure}[t]{0.49\textwidth}
         \centering
         \includegraphics[width=\textwidth]{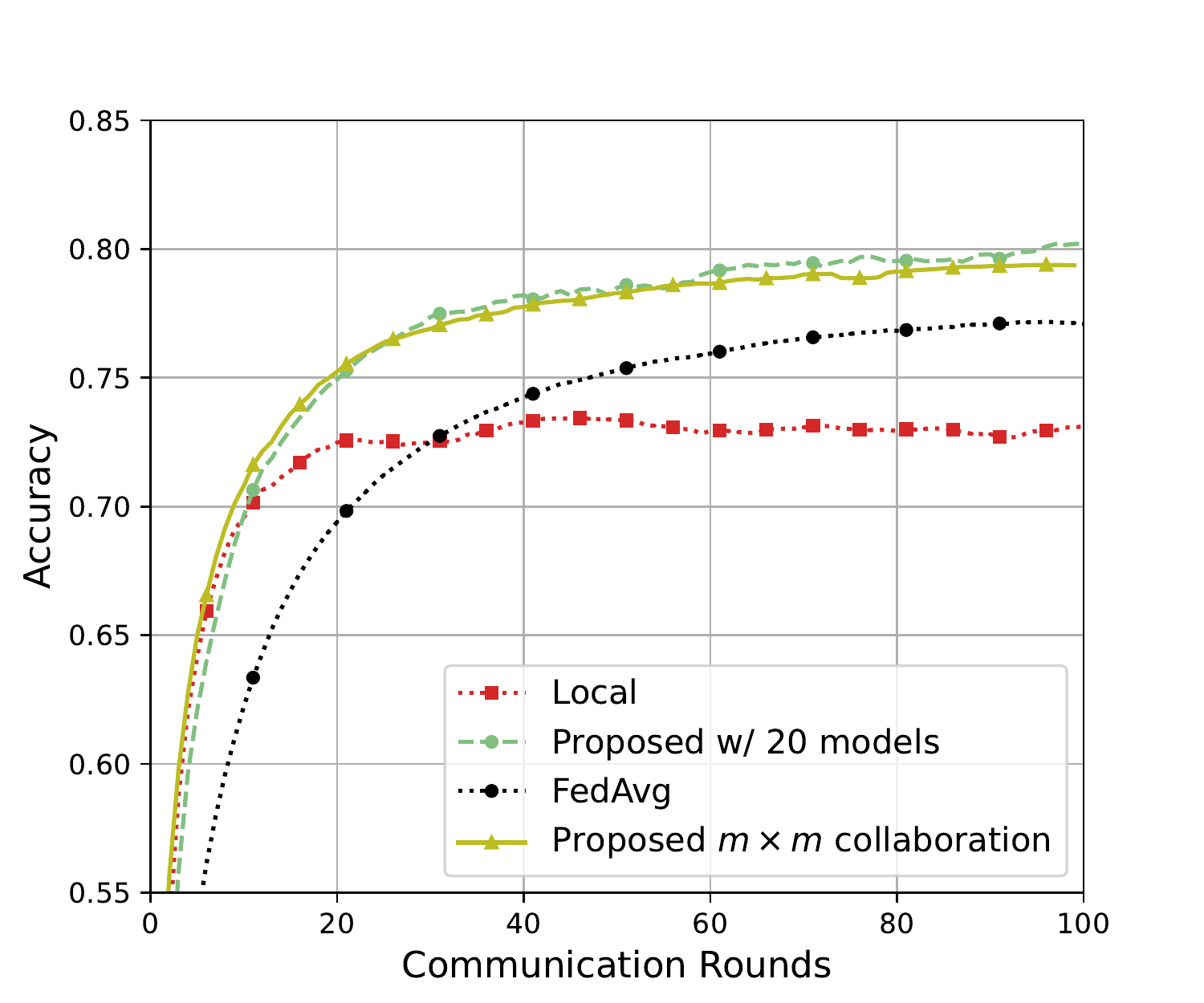}
         \centering
         \caption{\centering EMNIST label shift}
         \label{fig:1}
     \end{subfigure}
      \begin{subfigure}[t]{0.49\textwidth}
         \centering
         \includegraphics[width=\textwidth]{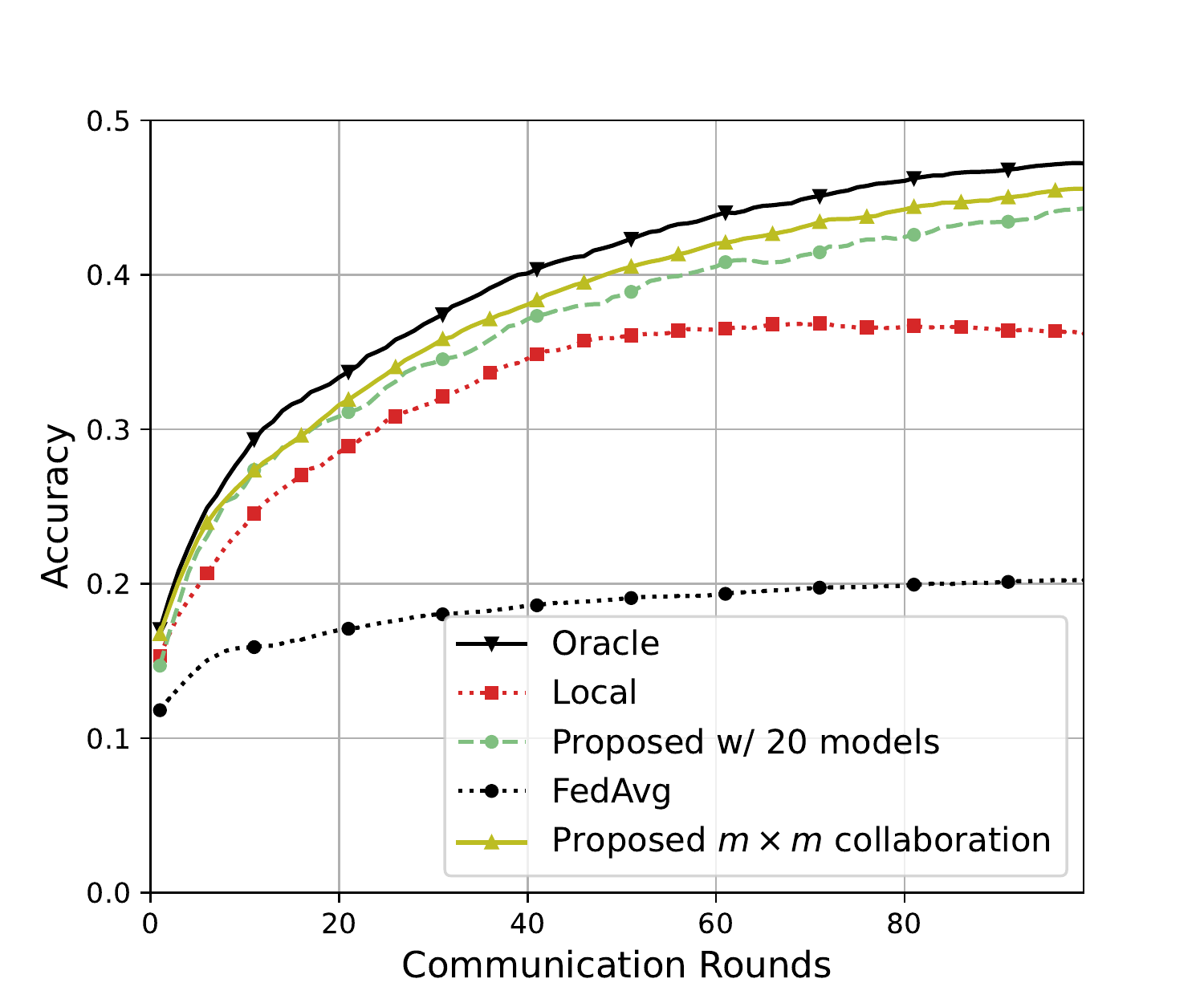}
         \caption{\centering CIFAR10 label \& covariate shift}
         \label{fig:2}
     \end{subfigure}
     \caption{Comparison between the proposed algorithm and the parallel user centric federated learning approach. The validation accuracy is averaged over 5 experiment runs.}
     \label{fig:nxn}
\end{figure*}

\subsection{Comparison with Parallel User-centric FL}
Even if the proposed user-centric aggregation rules outperform state-of-the-art personalized FL approaches, the resulting optimization procedure departs from the standard FL in the following sense: In the typical FL framework, at each communication round $t$, the PS aggregates the models that were locally trained, at each participating device, starting from the same launch model $\theta^{t-1}$. On the contrary, according to our proposed framework, devices may optimize different models depending on the specific user-centric aggregation rule they have been assigned to. This design choice is motivated by the assumption that the models of statistically similar propagate towards the same neighbourhood of the parameter space during the optimization \cite{zhang2020personalized}. As a result, in the proposed aggregation rule, models that are largely weighted, therefore associated with similar users, were locally optimized starting from similar initial parameters. Furthermore, if we were to adhere to the traditional FL procedure, and produce an exact minimizer of (\ref{centricloss}), we would have to run in parallel as many FL instances as the number of personalized streams $m_t$ and incur a $m_t$-fold computation and uplink communication load. 

To assess the quality of our assumption, we consider running in parallel $m_t$ collaborative FL instances employing the proposed user-centric weights and solving exactly $(\ref{centricloss})$ for each different aggregation rule. At each communication round, each user also optimizes the user-centric models of the other $m_t-1$ personalized streams which are then used at the PS server to apply the user-centric aggregation
rules \begin{equation}
\begin{aligned}
\centering 
\theta^{{t}}_i \leftarrow \sum^m_{j=1}w_{i,j}\theta^{{t-1/2}}_{i,j} \hspace{0.7cm} \text{for } i=1,2\cdots,m.
\end{aligned}
\label{eq13}
\end{equation}
Note that the aggregation rule in (\ref{eq13}) is different from the one in (\ref{eq2}), as $\theta^{{t-1/2}}_{i,j}$ denotes the update of user $j$ to the model of user $i$ obtained by locally optimizing $\theta_j^{t-1}$. 

We experiment using the EMNIST data set with label shift and the CIFAR10 data set with covariate and label shift. We set $m_t=m=20$ and use the same neural network model and settings indicated in Sec. \ref{section5}. In Fig. \ref{fig:nxn}, we report the performance of the parallel collaborative FL approach compared to our personalization strategy. For reference, we also report the performance of the FedAvg, local learning and oracle baselines. First, we notice that the fully collaborative solution performance serves as an upper bound to our personalization approach and that the oracle slightly outperforms the fully collaborative approach, which highlights the sub-optimality of our heuristic weighting scheme. However, the slight performance gain of the fully collaborative approach compared to our personalization strategy comes at the expense of $m_t$ times larger uplink communication load and computation cost at each edge device. These empirical results support our assumption: Even if the updated models are trained starting from different points in the parameter space at each communication round, the user-centric weighting scheme can direct statistically similar models in a neighbourhood across the loss landscape during training.
\subsection{Variance Computation: Mini-batch Size}

\begin{figure}[h]
\centering

         \includegraphics[width=0.45\textwidth]{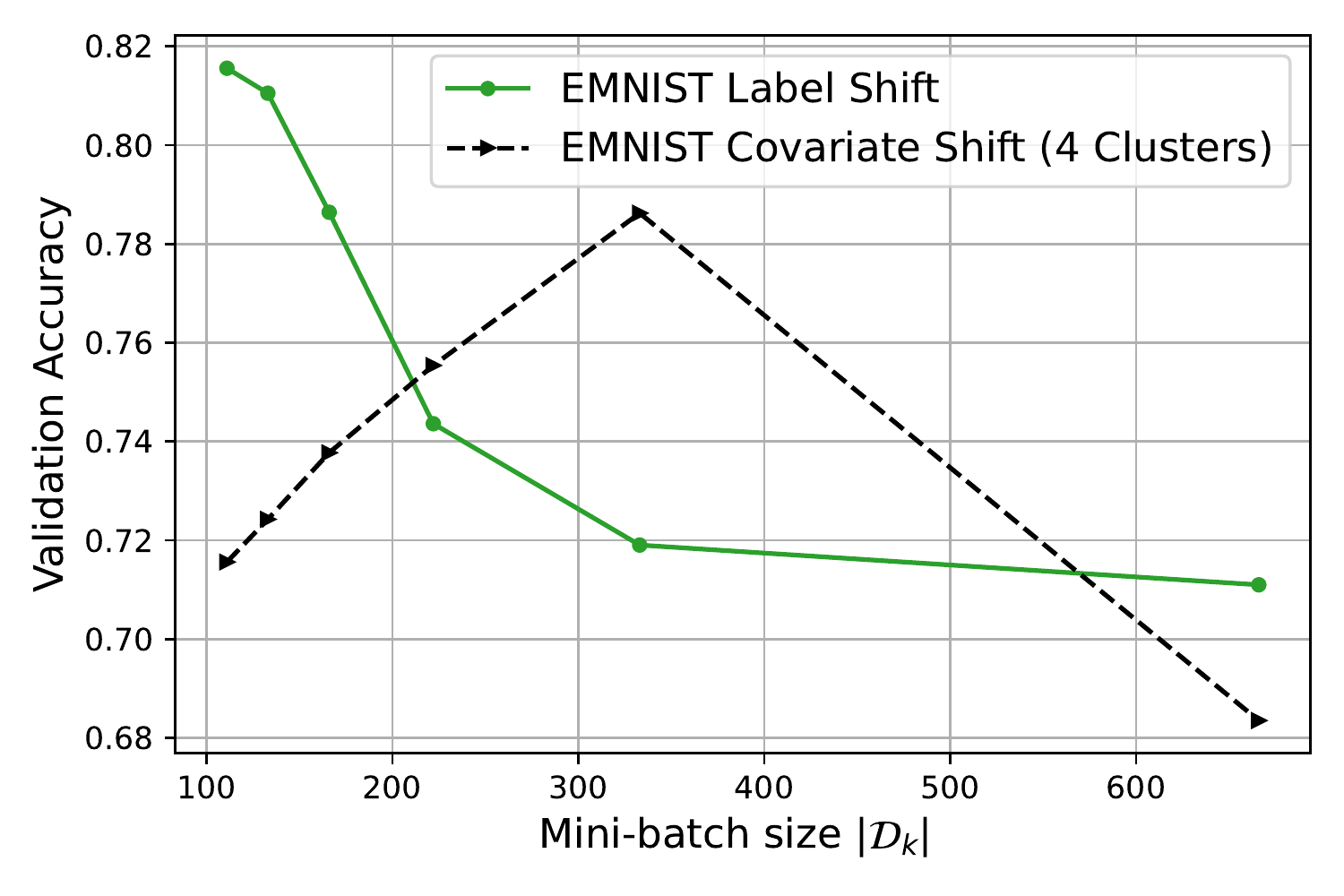}
         \centering
         \caption{\centering\small Effect of the mini-batch sizes on the maximum validation accuracy attained: A proxy to the quality of the calculated collaboration coefficients}
         \label{fig:var}
\end{figure}

As mentioned in section \ref{collaboration}, the mini-batch sizes chosen to calculate the variances play an essential role in the quality of the derived weights, i.e their ability to couple statistically similar users in the federated system. In Fig. \ref{fig:var}, we report the validation accuracy attained in an EMNIST label shift and covariate shift experiments. In both experiments, we randomly split 100k EMNIST data points across 100 users, i.e 1000 samples per user. Heterogeneity is introduced in both settings akin to the "label shift", and "label and covariate shift" settings in section \ref{setup}, respectively. We vary the mini-batch sizes used to calculate the variances from $ 100 \longrightarrow 660$ samples to explore the effect of this parameter on the validation accuracy of our personalization strategy in both scenarios. First, we note that according to (\ref{eq9}), decreasing the mini-batch size would yield an increase in the variance value as a result of the noisy gradients obtained compared to the average gradient computed over each user data set. In this case, our proposed aggregation rule renders similar to FedAvg, enabling collaboration among all users in the federated system, while still managing to softly couple statistically similar users under the assumption that $ \E_{\mathcal{D}_i,\mathcal{D}_j\sim P_i}\left[\Delta_{i,j}\right] \le  \E_{\mathcal{D}_i\sim P_i, \mathcal{D}_k\sim P_k} \left[\Delta_{i,k}\right]$ given that $ d_\mathcal{F}(P_i,P_k) > 0 $. This condition is favourable in the label shift setting while being detrimental to the extremely heterogeneous co-variate shift experiment, as it enables collaboration among users with competing tasks. Our claim is verified by the performance attained by our personalization rule in Fig. \ref{fig:var}, achieving a high validation accuracy in the label shift setting, while suffering in the co-variate shift experiment with a performance comparable to that of FedAvg attained in Fig. \ref{fig:EMNIST_cov} $(\sim 70.5 \,\%)$. However, as we increase the mini-batch size, the variances converge towards zero and our personalization algorithm degenerates to local training which is detrimental to both settings. Therefore, we conclude that the mini-batch size can be seen as a hyper-parameter for our algorithm, to be tuned according to the local data set size and the type of heterogeneity present across the learners. In our experiments presented in Fig. \ref{acc}, we set the mini-batch size $n_k = 100$ for the label shift experiment, and $n_k = N/3$ for the other two EMNIST co-variate and CIFAR10 concept shift experiments, where $N$ denotes the local data set size of each user.

\section{Conclusion}
\label{section6}
In this work, we have presented a novel FL personalization framework that exploits multiple user-centric aggregation rules to produce personalized models. The aggregation rules are based on user-specific mixture coefficients that can be computed during one communication round prior to federated training and are designed based on an excess risk upper bound of the weighted aggregated loss minimizer. Additionally, in order to limit the communication burden of personalization, we have proposed a $K$-means clustering algorithm to lump together users based on their similarity and serve each group of similar users with a single personalized model. In order to effectively trade communication resources for personalization capabilities, we have proposed to use the silhouette score to tune the number of user-centric aggregation rules at the PS before training commences. We have studied the performance of the proposed solution across different tasks. Overall, our solution yields personalized models with higher testing accuracy while at the same time being more communication-efficient compared to other state-of-the-art personalized FL baselines. 


\section{Appendix}
\subsection*{Proof of Theorem \ref{Th1}}
\label{A}
Denote by $f^*$ the $\argmin_{f\in \mathcal{F}} E_{z\sim P_i}[\ell(f,z)]$ and bound the estimation error of $\hat{f}_{\vec{w}_i}$ as 
\begin{align*}
    &Exc(\hat{f}_{\vec{w}_i},P_i)=E_{z\sim P_i}[\ell(\hat{f}_{\vec{w}_i},z)]-E_{z\sim P_i}[\ell(f^*,z)]\\
    &\leq E_{z\sim P_{\vec{w}_i}}[\ell(\hat{f}_{\vec{w}_i},z)]-E_{z\sim P_{\vec{w}_i}}[\ell(f^*,z)] + 2d_{\mathcal{F}}(P_i,P_{\vec{w}_i})+2\lambda\\
    &\leq E_{z\sim P_{\vec{w}_i}}[\ell(\hat{f}_{\vec{w}_i},z)]-\inf_{f\in \mathcal{F}}E_{z\sim P_{\vec{w}_i}}[\ell(f,z)] \\&+2 \sum_{j=1}^m w_{i,j}d_{\mathcal{F}}(P_i,P_{j})+2\lambda
\end{align*}
where $\lambda=\argmin_{f\in \mathcal{F}}\left( E_{z\sim P_i}[\ell(f,z)]+E_{z\sim P_{\vec{w}_i}}[\ell(f,z)]\right)$. We recognize the estimation error of $\hat{f}_{\vec{w}_i}$ w.r.t to the measure $P_{\vec{w}_i}$ that can be bounded following fairly standard approaches. In particular, 
\[
 E_{z\sim P_{\vec{w}_i}}[\ell(\hat{f}_{\vec{w}_i},z)]-\inf_{f\in \mathcal{F}}E_{z\sim P_{\vec{w}_i}}[\ell(f,z)]\leq 2 \Delta(\mathcal{G},Z)
\]
where 
\[\Delta(\mathcal{G},Z)=\sup_{g\in \mathcal{G}} \left|E_{P_{\vec{w}_i}}[g(Z)]-\sum^m_{j=1}\frac{w_{i,j}}{n_i} \sum_{z\in \mathcal{D}_i}g(z)\right|.
\] 
is the uniform deviation term and
\begin{equation*}
    \mathcal{G}=\left\{Z \xrightarrow{} \ell(f,Z): f \in \mathcal{F}\right\}.
\end{equation*}
is the class resulting from the composition of the loss function $\ell(\cdot)$ and $\mathcal{F}$.
The uniform deviation bound can be bounded in different ways, depending on the type of knowledge about the random variable $g(Z)$, in the following we assume that the loss function is bounded with range $B$ and we exploit Azuma's inequality. In particular, the Doob's Martingale associated to the weighted loss will still have increments bounded by $\frac{w_{i,j}}{n_i} B$ depending to which loss term the increment is associated. Recognizing this, we can then directly apply Azuma's concentration bound and state that w.p. $1-\delta$ the following holds
\[
\Delta(\mathcal{G},Z)\leq E_P[\Delta(\mathcal{G},Z)]+B\sqrt{\sum^m_{j=1}\frac{w_{i,j}^2}{n_j}\log\left(\frac{2}{\delta}\right)}
\]
Finally, the expected uniform deviation can be bounded by the Rademacher complexity as follows
\[
 E_P[\Delta(\mathcal{G},Z)]\leq2\text{Rad}(\mathcal{G})
\]
where 
\[
\text{Rad}(\mathcal{G})=E_{\vec{\sigma},\mathcal{D}_1,\dots,\mathcal{D}_j}\left[\sup_{g\in \mathcal{G}} \sum^m_{j=1}\frac{w_{i,j}}{n_i}\sum_{i=1}^{n_i}\sigma_{i,j} g(Z_{i,j})\right]
\]
By a direct application of Massart's and Sauer's  Lemma we obtain
\begin{align*}
&\text{Rad}(\mathcal{G})\leq \sqrt{\sum^m_{j=1}\frac{w_{i,j}^2}{n_j}}\\&\times\sqrt{\frac{2\text{VCdim}(\mathcal{G})\left( \log(e\sum_j n_j)+\log(\text{VCdim}(\mathcal{G}))\right)}{\sum_j n_j}}
\end{align*}
combining everything together, we get the final result.
\subsection*{Proof of Theorem \ref{Th2}}
Thanks to the upper bound on the target domain risk and the fact that the sum of two sub-Gaussian random variables of parameter $\sigma$ is also sub-Gaussian with parameter $2\sigma$, we can decompose the excess risk as 
\begin{align*}
    &Exc(\hat{f}_{\vec{w}_i},P_i)=E_{z\sim P_i}[\ell(\hat{f}_{\vec{w}_i},z)]-\inf_{f\in \mathcal{F}}E_{z\sim P_i}[\ell(f,z)]\\
   &=E_{z\sim P_i}[\ell(\hat{f}_{\vec{w}_i},z)-\ell(f^*,z)]\\
   &\leq E_{z\sim P_{\vec{w}_i}}[\ell(\hat{f}_{\vec{w}_i},z)-\ell(f^*,z)]+2\beta\sigma^2+\frac{D_{JS}(P_i||P_{\vec{w}_i})}{\beta}
\end{align*}
From the convexity of the KL-divergence we can bound the Jensen-Shannon divergence as follows
{\small 
\begin{align*}
    &D_{JS}(P_i||P_{\vec{w}_i}) =\frac{1}{2}KL\left(P_i||\frac{P_i+P_{\vec{w}_i}}{2}\right)+\frac{1}{2}KL\left(P_{\vec{w}_i}||\frac{P_i+P_{\vec{w}_i}}{2}\right)\\
    &=\frac{1}{2}KL\left(P_i||\frac{\sum_jw_{i,j}(P_i+P_j)}{2}\right)\\&+\frac{1}{2}KL\left(\sum_jw_{i,j}P_j||\frac{\sum_jw_{i,j}(P_i+P_j)}{2}\right)\\&\leq
    \hspace{0.4cm}\frac{1}{2}\sum_j w_{i,j} \left( KL\left(P_i||\frac{(P_i+P_j)}{2}\right)+KL\left(P_j||\frac{(P_i+P_j)}{2}\right)\right)\\
    &=\sum_j w_{i,j}D_{JS}(P_i||P_j)
\end{align*}
Plugging it back into the previous expression and minimizing with respect to $\beta$ we obtain
\begin{align*}
  Exc(\hat{f}_{\vec{w}_i},P_i) &\leq E_{z\sim P_{\vec{w}_i}}[\ell(\hat{f}_{\vec{w}_i},z)]-\inf_{f\in \mathcal{F}}E_{z\sim P_{\vec{w}_i}}[\ell(f,z)]\\&+2\beta\sigma^2+\frac{\sum_j \vec{w}_{i,j}D_{JS}(P_i||P_j)}{\beta} \\
   &\leq E_{z\sim P_{\vec{w}_i}}[\ell(\hat{f}_{\vec{w}_i},z)]-\inf_{f\in \mathcal{F}}E_{z\sim P_{\vec{w}_i}}[\ell(f,z)]\\&+2\sigma\sqrt{2\sum^m_{j=1}D_{JS}(P_i||P_j)} \\
\end{align*}}%
We identify the estimation error and we bound as previously done for Theorem $\ref{Th1}$ to obtain the final result. Moreover, for $B$-bounded random variables, $\sigma=B/2$

 

        

\bibliographystyle{unsrt}
\bibliography{biblio}

\begin{thebibliography}{10}

\bibitem{health}
Jemal~H. Abawajy and Mohammad~Mehedi Hassan.
\newblock Federated internet of things and cloud computing pervasive patient
  health monitoring system.
\newblock {\em IEEE Communications Magazine}, 55(1):48--53, 2017.

\bibitem{mcmahan2017communication}
Brendan McMahan, Eider Moore, Daniel Ramage, Seth Hampson, and Blaise~Aguera
  y~Arcas.
\newblock Communication-efficient learning of deep networks from decentralized
  data.
\newblock In {\em Artificial Intelligence and Statistics}, pages 1273--1282.
  PMLR, 2017.

\bibitem{sattler2020clustered}
Felix Sattler, Klaus-Robert M{\"u}ller, and Wojciech Samek.
\newblock Clustered federated learning: Model-agnostic distributed multitask
  optimization under privacy constraints.
\newblock {\em IEEE Transactions on Neural Networks and Learning Systems},
  2020.

\bibitem{li2018federated}
Tian Li, Anit~Kumar Sahu, Manzil Zaheer, Maziar Sanjabi, Ameet Talwalkar, and
  Virginia Smith.
\newblock Federated optimization in heterogeneous networks.
\newblock {\em arXiv preprint arXiv:1812.06127}, 2018.

\bibitem{prevwork}
Mohamad Mestoukirdi, Matteo Zecchin, David Gesbert, Qianrui Li, and Nicolas
  Gresset.
\newblock {User-Centric} federated learning.
\newblock In {\em 2021 IEEE Globecom Workshops (GC Wkshps): Workshop on
  Wireless communications for distributed intelligence (GC 2021 Workshop -
  WCDI)}, Madrid, Spain, December 2021.

\bibitem{briggs2020federated}
Christopher Briggs, Zhong Fan, and Peter Andras.
\newblock Federated learning with hierarchical clustering of local updates to
  improve training on non-iid data.
\newblock In {\em 2020 International Joint Conference on Neural Networks
  (IJCNN)}, pages 1--9. IEEE, 2020.

\bibitem{zhang2020personalized}
Michael Zhang, Karan Sapra, Sanja Fidler, Serena Yeung, and Jose~M Alvarez.
\newblock Personalized federated learning with first order model optimization.
\newblock {\em arXiv preprint arXiv:2012.08565}, 2020.

\bibitem{mixture2021s}
Othmane Marfoq, Giovanni Neglia, Aurélien Bellet, Laetitia Kameni, and Richard
  Vidal.
\newblock Federated multi-task learning under a mixture of distributions.
\newblock {\em International Workshop on Federated Learning for User Privacy
  and Data Confidentiality in conjunction with ICML 2021 (FL-ICML'21)}, 2021.

\bibitem{reisser2021federated}
Matthias Reisser, Christos Louizos, Efstratios Gavves, and Max Welling.
\newblock Federated mixture of experts.
\newblock {\em arXiv preprint arXiv:2107.06724}, 2021.

\bibitem{DBLP:journals/corr/abs-1812-06127}
Anit~Kumar Sahu, Tian Li, Maziar Sanjabi, Manzil Zaheer, Ameet Talwalkar, and
  Virginia Smith.
\newblock On the convergence of federated optimization in heterogeneous
  networks.
\newblock {\em CoRR}, abs/1812.06127, 2018.

\bibitem{DBLP:journals/corr/abs-1910-06378}
Sai~Praneeth Karimireddy, Satyen Kale, Mehryar Mohri, Sashank~J. Reddi,
  Sebastian~U. Stich, and Ananda~Theertha Suresh.
\newblock {SCAFFOLD:} stochastic controlled averaging for on-device federated
  learning.
\newblock {\em CoRR}, abs/1910.06378, 2019.

\bibitem{DBLP:journals/corr/abs-2012-04221}
Tian Li, Shengyuan Hu, Ahmad Beirami, and Virginia Smith.
\newblock Ditto: Fair and robust federated learning through personalization.
\newblock In Marina Meila and Tong Zhang, editors, {\em Proceedings of the 38th
  International Conference on Machine Learning}, volume 139 of {\em Proceedings
  of Machine Learning Research}, pages 6357--6368. PMLR, 18--24 Jul 2021.

\bibitem{DBLP:journals/corr/abs-2006-08848}
Canh~T. Dinh, Nguyen~H. Tran, and Tuan~Dung Nguyen.
\newblock Personalized federated learning with moreau envelopes.
\newblock {\em CoRR}, abs/2006.08848, 2020.

\bibitem{ben2010theory}
Shai Ben-David, John Blitzer, Koby Crammer, Alex Kulesza, Fernando Pereira, and
  Jennifer~Wortman Vaughan.
\newblock A theory of learning from different domains.
\newblock {\em Machine learning}, 79(1):151--175, 2010.

\bibitem{mansour2009domain}
Yishay Mansour, Mehryar Mohri, and Afshin Rostamizadeh.
\newblock Domain adaptation: Learning bounds and algorithms.
\newblock {\em arXiv preprint arXiv:0902.3430}, 2009.

\bibitem{shui2020beyond}
Changjian Shui, Qi~Chen, Jun Wen, Fan Zhou, Christian Gagn{\'e}, and Boyu Wang.
\newblock Beyond h-divergence: Domain adaptation theory with jensen-shannon
  divergence.
\newblock 2020.

\bibitem{cohen2017emnist}
Gregory Cohen, Saeed Afshar, Jonathan Tapson, and Andre Van~Schaik.
\newblock Emnist: Extending mnist to handwritten letters.
\newblock In {\em 2017 International Joint Conference on Neural Networks
  (IJCNN)}, pages 2921--2926. IEEE, 2017.

\bibitem{krizhevsky2009learning}
Alex Krizhevsky, Geoffrey Hinton, et~al.
\newblock Learning multiple layers of features from tiny images.
\newblock 2009.

\bibitem{wang2020tackling}
Jianyu Wang, Qinghua Liu, Hao Liang, Gauri Joshi, and H~Vincent Poor.
\newblock Tackling the objective inconsistency problem in heterogeneous
  federated optimization.
\newblock {\em arXiv preprint arXiv:2007.07481}, 2020.

\bibitem{lecun1998gradient}
Yann LeCun, L{\'e}on Bottou, Yoshua Bengio, and Patrick Haffner.
\newblock Gradient-based learning applied to document recognition.
\newblock {\em Proceedings of the IEEE}, 86(11):2278--2324, 1998.

\end{thebibliography}

\end{document}